\pdfoutput=1
\PassOptionsToPackage{table}{xcolor}
\documentclass[10pt,twocolumn,letterpaper]{article}

\usepackage[pagenumbers]{iccv} %

\usepackage{subcaption,booktabs}
\usepackage[table]{xcolor}
\definecolor{lightcyan}{rgb}{0.88,1,1}
\usepackage{pifont}
\usepackage[utf8]{inputenc} %
\usepackage[T1]{fontenc}    %
\usepackage{url}            %
\usepackage{booktabs}       %
\usepackage{amsfonts}       %
\usepackage{nicefrac}       %
\usepackage{microtype}      %
\usepackage{graphicx}
\usepackage{multirow}
\usepackage{amsmath}
\usepackage{amssymb}
\usepackage{colortbl} 
\usepackage{fontawesome5}
\usepackage{bbm}
\usepackage{comment}
\usepackage{cuted}
\usepackage{capt-of}
\usepackage[skip=1pt]{parskip}

\usepackage[pagebackref,breaklinks,colorlinks,citecolor=blue, linkcolor=blue]{hyperref}
\newcommand{\methodname}{\texttt{MoSiC}\xspace}

\def\paperID{13782} %
\def\confName{ICCV}
\def\confYear{2025}

\title{\methodname: Optimal-Transport Motion Trajectory \\ for Dense Self-Supervised Learning}
\vspace{-3pt}
\author{Mohammadreza Salehi,$^1$\thanks{Equal Contribution. Correspondence: s.salehidehnavi@uva.nl
}\hspace{0.5em} Shashanka Venkataramanan,$^{2\ast}$ \hspace{0.5em} Ioana Simion,$^1$ \hspace{0.5em} \\
Efstratios Gavves,$^1$ \hspace{0.5em} Cees G. M. Snoek,$^1$ \hspace{0.5em} Yuki M Asano$^3$ \\
$^1$ VIS Lab, UvA \hspace{0.5em} $^2$ Valeo.ai \hspace{0.5em} $^3$ Fundamental AI Lab, UTN
}

\definecolor{LightCyan}{rgb}{0.88,1,1}
\definecolor{ForestGreen}{RGB}{34,139,34}

\newcommand{\fig}[2][1]{\includegraphics[width=#1\linewidth]{figs/#2}}

\newcommand{\real}{\mathbb{R}}

\begin{document}
\maketitle

\begin{abstract}
Dense self-supervised learning has shown great promise for learning pixel- and patch-level representations, but extending it to videos remains challenging due to the complexity of motion dynamics. Existing approaches struggle as they rely on static augmentations that fail under object deformations, occlusions, and camera movement, leading to inconsistent feature learning over time.
We propose a motion-guided self-supervised learning framework that clusters dense point tracks to learn spatiotemporally consistent representations. By leveraging an off-the-shelf point tracker, we extract long-range motion trajectories and optimize feature clustering through a momentum-encoder-based optimal transport mechanism. To ensure temporal coherence, we propagate cluster assignments along tracked points, enforcing feature consistency across views despite viewpoint changes. Integrating motion as an implicit supervisory signal, our method learns representations that generalize across frames, improving robustness in dynamic scenes and challenging occlusion scenarios.
By initializing from strong image-pretrained models and leveraging video data for training, we improve state-of-the-art by 1\% to 6\% on six image and video datasets and four evaluation benchmarks. The implementation is publicly available at our GitHub repository: \href{https://github.com/SMSD75/MoSiC/tree/main}{\texttt{github.com/SMSD75/MoSiC}}
\end{abstract}
\vspace{-6pt}

\section{Introduction}
\label{sec:intro}  

Dense self-supervised learning has emerged as a powerful paradigm for learning rich pixel- and patch-level representations without reliance on labeled data~\cite{ziegler2022self, henaff2021efficient, van2021unsupervised}. While significant progress has been made in the image domain, leveraging videos for self-supervised learning presents an even greater opportunity. Videos not only provide vast amounts of readily available data but also introduce a natural temporal dimension, which could be crucial for learning representations that extend beyond static imagery. However, directly applying image-based self-supervised learning techniques to videos proves ineffective~\cite{tong2022videomae, assran2023self}. In images, correspondences are often implicitly encoded by data augmentations—for instance, color transformations preserve pixel-wise associations. In contrast, videos exhibit complex motion patterns caused by object movement, camera shifts, and dynamic backgrounds, breaking such pixel-level correspondences.

\begin{figure}[t]
    \centering
    \includegraphics[width=1.\linewidth]{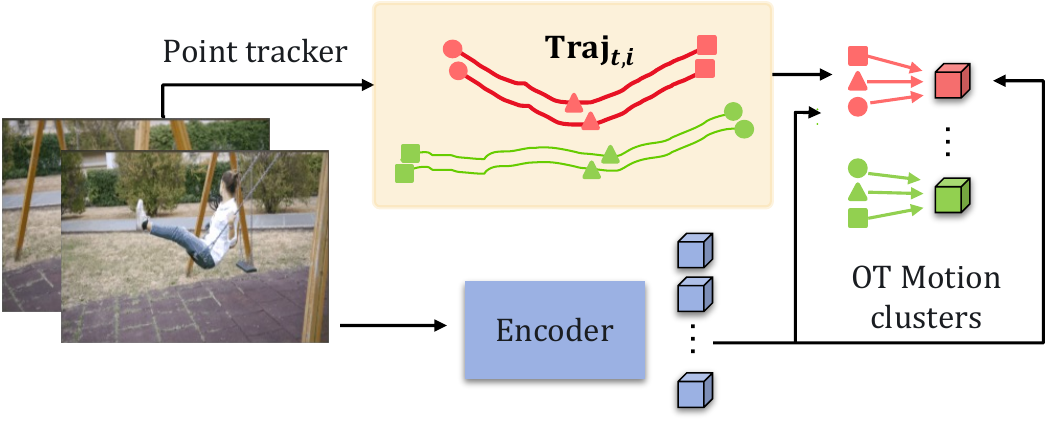}
    \vspace{-6pt}
    \caption{We introduce \methodname, a self-supervised framework that uses optimal transport (OT) to enforce spatio-temporal consistency in dense visual representations. By tracking points across frames and clustering their motion trajectories, \methodname encourages features of points that move together to be temporally coherent.} 
    \vspace{-6pt}
    \label{fig:teaser}
\end{figure}

\quad In this work, we aim to learn temporally coherent dense representations by leveraging motion as an implicit supervisory signal. Rather than treating frames independently, we enforce feature consistency over time by clustering points that move together. This encourages representations that are not only spatially meaningful but also temporally stable, improving robustness to occlusions and viewpoint changes. Prior works have explored various strategies for learning dense representations in videos, including architectural modifications to track multiple objects in ViTs~\cite{venkataramanan2023imagenet}, joint-embedding architectures with specialized masking strategies~\cite{assran2023self, bardes2024revisiting}, and temporally coherent clustering~\cite{salehi2023time}. In this work, we propose a new approach inspired by the Gestalt principle of "\emph{what moves together belongs together}"~\cite{koffka2013principles}. We bring this principle to a much finer level of granularity and introduce~\methodname in~\autoref{fig:teaser}, a self-supervised \textit{Mo}tion-based \textit{Si}nkhorn \textit{C}lustering framework that clusters \emph{points} along motion tracks to learn spatiotemporally consistent representations.

\begin{figure}[t]
    \centering
    \includegraphics[width=1.\linewidth]{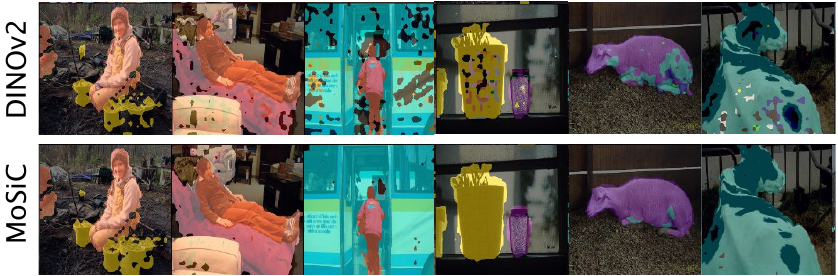}
    \vspace{-6pt}
    \caption{\textbf{DINOv2 vs. \methodname in-context scene understanding qualitative comparison on Pascal VOC.} \methodname trains DINOv2’s dense representations on unlabeled videos and improves them, resulting in more precise segmentation boundaries and better object identification.} 
    \label{fig:dinov2_vs_ours}
\end{figure}

Our approach begins by extracting dense point tracks that capture long-range motion trajectories across frames. These tracked points are then clustered using a momentum-encoder-based optimal transport mechanism, ensuring feature consistency while applying the loss only to visible points. By enforcing temporal coherence along motion trajectories, our model captures object permanence despite occlusions and viewpoint changes, resulting in more stable and semantically meaningful representations. Extensive evaluations across six image and video datasets demonstrate the effectiveness of~\methodname, showcasing its ability to learn robust and transferable dense features for tasks such as semantic segmentation and scene understanding. Notably, by leveraging the strong vision foundation model DINOv2~\cite{oquab2023dinov2}, \methodname consistently improves its dense representations, as demonstrated qualitatively in \autoref{fig:dinov2_vs_ours}. This improvement is further validated quantitatively, achieving a 1\% to 6\% gain across four benchmarks—solely through training on video data.

\section{Related Works}
\label{sec:related}
\paragraph{Video-to-Image self-supervised learning}
Video data provides rich temporal signals that naturally supervise learning by capturing object transformations and scene dynamics over time. Unlike static images, videos encode motion patterns and object persistence, offering essential cues for robust visual understanding. Early approaches explored various pretext tasks, including egomotion prediction~\cite{agrawal2015learning, jayaraman2015learning}, pose estimation~\cite{chakraborty2017learning}, dense prediction~\cite{pathak2017learning, li2019joint}, optical flow~\cite{mahendran2019cross, xiong2021self}, and visual correspondence learning~\cite{wang2019learning}.

Traditional self-supervised learning (SSL) methods, primarily developed on ImageNet, rely on augmentations for contrastive or predictive objectives~\cite{chen2020simple, caron2021emerging}. While extended to video~\cite{parthasarathy2022self, gordon2020watching, tschannen2020self, wang2015unsupervised}, direct transfer often leads to performance drops due to distribution shifts~\cite{parthasarathy2022self}. Recent works instead learn directly from videos, avoiding augmentation-based supervision. Carreira \etal~\cite{carreira2024learning} process continuous video streams without mini-batches, while DoRA~\cite{venkataramanan2023imagenet} uses a single, long, unlabelled video to train strong image encoders without relying on traditional data augmentation techniques. MooG~\cite{van2025moving} models dynamic scene elements for improved motion tracking, and \cite{carreira2024scaling} scales SSL to larger models via masked autoencoding for spatiotemporal learning.

Closest to our work, TimeTuning~\cite{salehi2023time} leverages temporal cues but struggles with occlusions and long-range tracking under rapid camera motion. Unlike existing SSL methods, we incorporate dense point tracking to ensure robust temporal correspondences, improving performance on real-world video and image datasets.

\paragraph{Unsupervised object segmentation}
Unsupervised object segmentation has been widely studied, with many methods~\cite{bao2023object, lan2023smooseg, araslanov2021dense, zadaianchuk2023object, hamilton2022unsupervised, simeoni2023unsupervised} segmenting objects in images without labeled data. These approaches often use pretrained models to extract information and train another model for segmentation. Seitzer \etal~\cite{seitzer2022bridging} use slot attention to reconstruct DINO-pretrained features, clustering image regions into object slots. CroC~\cite{stegmuller2023croc} aligns cluster centers across views using a dense self-supervised loss, while Leopart~\cite{ziegler2022self} enhances representations with dense clustering. Hummingbird~\cite{balazevic2023towards} leverages attention mechanisms within and across images for in-context scene understanding. CrIBo~\cite{lebailly2023cribo} enforces cross-image nearest-neighbor consistency, while NeCo~\cite{pariza2025near} enforces patch-level nearest-neighbor consistency generating high-quality dense representations.

While these methods demonstrate the effectiveness of clustering and feature alignment for object segmentation, they primarily focus on static images, lacking the temporal consistency crucial for videos. In contrast, our method extends these ideas to the video domain by leveraging motion trajectories as signal. By propagating cluster assignments along tracked points, we ensure spatiotemporally coherent representations, bridging the gap between object-centric clustering and motion-aware learning. Tracktention~\cite{lai2025tracktention} is another work that uses point trackers, which guides the attention for video prediction tasks to improve temporal consistency for depth. In contrast, we use point tracks to enforce temporal consistency for object segmentation across time.

\section{\methodname: Motion-Based Sinkhorn Clustering}
\label{sec:method}

\begin{figure*}
\centering
\fig[1]{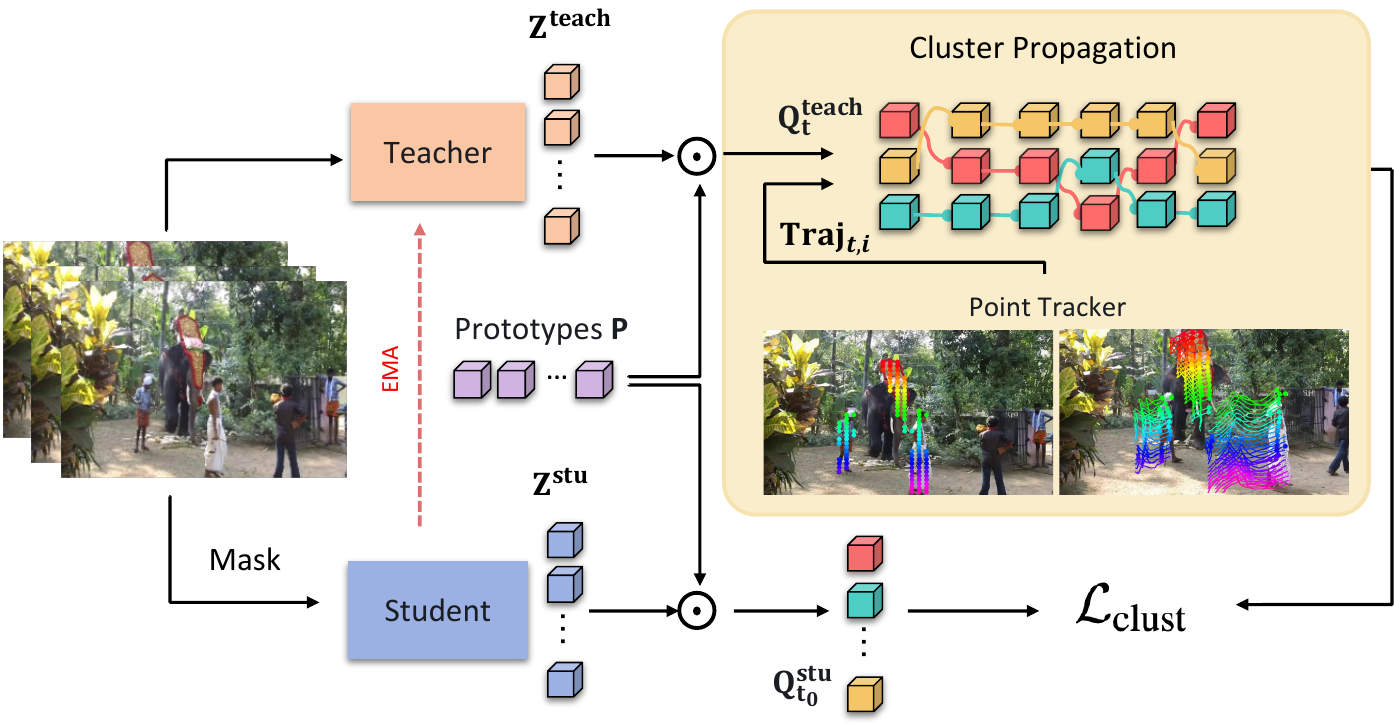}
\caption{\textbf{\methodname}: Motion-based Sinkhorn Clustering for dense self-supervised pretraining. A video clip is patchified, with masked patches passed through the student network to obtain $Z^{\text{stu}} \in \real^{n_s \times d \times T}$, while the teacher network processes the patches from raw clip to produce $Z^{\text{teach}} \in \real^{n \times d \times T}$. Features are clustered using Sinkhorn-Knopp~\cite{cuturi2013sinkhorn} into prototypes $P$. The cluster assignments $Q^{\text{teach}}_{t_0}$ (eq. \eqref{eq:stu-assign}) and $Q^{\text{stu}}_{t_0}$ (eq. \eqref{eq:teach-assign}) are computed for the first frame ($t=t_0$). These assignments are then propagated along motion trajectories $\text{Traj}_{t,i}$ computed using CoTracker-v3~\cite{karaev2024cotracker3} to obtain $Q^{\text{teach}}_{t}$ (eq. \eqref{eq:cotrack-clust-prop}). Finally, a cross-entropy loss is applied between $Q^{\text{stu}}_{t_0}$ and $Q^{\text{teach}}_{t}$ (eq. \eqref{eq:final-loss}) for frames where $\text{Traj}_{t,i}$ remains visible, ensuring temporal consistency by clustering points along motion paths and preserving object identity over time.}
\vspace{-6pt}
\label{fig:model}
\end{figure*}
We introduce \emph{\methodname} in~\autoref{fig:model}, a novel approach to dense self-supervised learning in videos that leverages motion as an implicit supervisory signal. Our method first extracts long-range motion trajectories using an off-the-shelf point tracker, ensuring robustness to occlusions and object permanence. To enforce spatiotemporal coherence, we perform clustering over these motion trajectories, aligning feature representations across frames. Unlike prior methods that rely purely on mask propagation or local frame-based learning, our approach establishes long-range correspondences, leading to spatiotemporally consistent features, considerably improving the quality of the learned representations.

\subsection{Preliminaries}
Given a video clip $X \in \real^{h \times w \times c \times T}$, where $T$ is the number of frames, $h \times w$ is the spatial resolution, and $c$ is the number of channels, we first divide each frame into $n = \frac{h \times w}{p^2}$ patches of size $p \times p$. These patches are then linearly projected into $d$-dimensional embeddings and passed through an encoder. We apply a binary mask $M \in \{0,1\}^{n \times T}$ uniformly across all frames, randomly masking a fraction $m$ of the tokens. Following Moutakanni \textit{et al.}~\cite{moutakanni2025you}, we demonstrate that simple augmentations, such as cropping and masking, are sufficient for learning effective representations, eliminating the need for more complex augmentations like color jitter, grayscale, or blurring. 
We use a Vision Transformer (ViT)~\cite{dosovitskiy2020image} in a teacher-student framework, where the student and teacher models are $f_{\theta}$ and $g_{\theta}$. The representations are given by $Z^{\text{teach}} = g_{\theta}(X_t) \in \mathbb{R}^{n \times d}$ for the teacher and $Z^{\text{stu}} = f_{\theta}((1-M) \odot X_t) \in \mathbb{R}^{n_s \times d}$ for the student networks, where $\odot$ denotes the Hadamard product and $n_s$ denotes the number of unmasked tokens. 
\subsection{Clustering Motion Trajectories}
A major challenge in learning from real videos is the transient nature of objects due to occlusions, camera movement, and object permanence. Unlike images, video frames undergo temporal deformations, making feature consistency difficult. Prior methods~\cite{salehi2023time} propagate object masks across frames but struggle when objects temporarily disappear, leading to propagation errors. Additionally, long-range tracking suffers from drift accumulation, which compounds over time and degrades feature representations—an issue that worsens with longer sequences. 
\vspace{-6pt}
\paragraph{Motion Trajectories From a Point Tracker}
To address these challenges, we utilize an off-the-shelf point tracker capable of long-range tracking and robust to object permanence. The tracker samples points from a regular grid in the first frame and updates their positions across subsequent frames. Given our video clip $X$, we sample $N$ points from the initial frame, forming a $\text{grid} = {(x_i, y_i)}_{i=1}^{N}$, where $(x_i, y_i)$ are the coordinates of the $i^{\text{th}}$ point. Using the video clip $X$ and the grid, the tracker predicts the trajectories of these points across all frames as:
\begin{equation}
\text{Traj}_{t,i} := \text{\texttt{Tracker}}(X_t, (x_i, y_i)) \in \real^{T \times N \times 2}.
\label{eq:traj}
\end{equation}
$\text{Traj}_{t,i}$ represents the trajectory of the $i^{\text{th}}$ point across $T$ frames, with each element encoding its coordinates at time $t$.
\vspace{-11pt}

\paragraph{Optimal-Transport Based Clustering}
Although motion trajectories provide a strong signal for motion-based grouping, relying solely on them can be a challenge especially for objects with subtle or ambiguous motion. To address this, we perform clustering to enforce temporal coherence, ensuring that points belonging to the same object or part remain consistently clustered over time. Given motion trajectories from the point tracker, we incorporate semantic clustering to obtain a more discriminative representation of objects.

We first cluster the features from the student and teacher networks $ Z^{\text{stu}} \in \real^{n_s \times d \times T} , Z^{\text{teach}} \in \real^{n \times d \times T} $ for the first frame using the Sinkhorn-Knopp algorithm~\cite{cuturi2013sinkhorn}. This clustering solves an entropy-regularized optimal transport problem, iteratively refining the assignment of feature tokens to cluster prototypes while maintaining marginal constraints and has been used in various self-supervised works~\cite{asano2019self, caron2020unsupervised, salehi2024sigma, oquab2023dinov2}.

Let $ P^{\text{stu}}, P^{\text{teach}} \in \real^{K \times d} $ denote $K$ cluster prototypes and $ Z^{\text{stu}}_{t_0} \in  \real^{n_s \times d} , Z^{\text{teach}}_{t_0} \in \real^{n \times d} $ represent the features extracted from the initial frame $t=t_0 $. The transport cost between the patch features and the cluster prototypes $ C^{\text{stu}} \in \real^{n_s \times K}$, $C^{\text{teach}} \in \real^{n \times K} $ are:
\begin{align}
C^{\text{stu}} &= - Z^{\text{stu}}_{t_0} P^{\text{stu}^\top} \in \real^{n_s \times K}  \label{eq:cost-stu}\\
C^{\text{teach}} &= - Z^{\text{teach}}_{t_0} P^{\text{teach}^\top} \in \real^{n \times K}  \label{eq:cost-teach} 
\end{align}
which denote the negative cosine similarity between the normalized patch features and normalized prototypes. The optimal transport plans $M^{\text{stu}*} $ and $ M^{\text{teach}*} $ using eq.~\eqref{eq:cost-stu} and eq.~\eqref{eq:cost-teach} are obtained by solving:
\begin{align}
M^{\text{stu}*} &= \arg\min_{M^\text{stu} \in \mathcal{M}^\text{stu}} \langle M^\text{stu}, C^\text{stu} \rangle - \epsilon\frac{1}{\lambda} H(M^\text{stu}) \label{eq:opt-stu}\\
M^{\text{teach}*} &= \arg\min_{M^\text{teach} \in \mathcal{M}^\text{teach}} \langle M^\text{teach}, C^\text{teach} \rangle - \epsilon\frac{1}{\lambda} H(M^\text{teach}) \label{eq:opt-teach}
\end{align}
where $M^\text{stu} \in \mathbb{R}^{n_s \times K}, M^\text{teach} \in \mathbb{R}^{n \times K} $ are the assignment matrices, $H(M)$ is an entropy regularization, and $\epsilon$ is the regularization coefficient than controls the smoothness of the assignment enforcing uniform marginal constraints 
$M^\text{stu} \mathbf{1}_K = \mathbf{1}_{n_s}/n_s, \quad M^{\text{stu}^\top} \mathbf{1}_{n_s} =  \mathbf{1}_K/K$ and 
\newline $M^\text{teach} \mathbf{1}_K =  \mathbf{1}_n /n, \quad M^{\text{teach}^\top} \mathbf{1}_n = \mathbf{1}_K/K$.

\subsection{Cluster Propagation using Motion Trajectories}
After clustering the initial frame, we propagate these assignments across frames using the tracked motion trajectories from eq.~\eqref{eq:traj}. This allows us to maintain consistent object representations over time, even when objects undergo deformation. Given the tracked trajectories from eq.~\eqref{eq:traj}, we sample features from the student and teacher embeddings $Z^{\text{stu}}, Z^{\text{teach}}$ at these locations.

To extract features at the tracked locations, we interpolate values from the feature maps using the continuous trajectory coordinates. Since these points do not always align with discrete pixel locations, we apply bilinear interpolation for the teacher network, computing feature values as a weighted sum of the four nearest neighbors. 
For the student network, we use nearest-neighbor interpolation, since the points are sampled along a simple uniformly spaced grid.
\vspace{-6pt}
\paragraph{Cluster Propagation}
First, we compute the cluster assignment for the features from the teacher and the student networks using $M^{\text{stu}*}$ and $M^{\text{teach}*}$ computed in eq.~\eqref{eq:opt-stu} and eq.~\eqref{eq:opt-teach} as follows:

\begin{align}
\mathcal{Q}^\text{stu}_{t_0} &= \arg\max_K M^\text{stu*} \cdot Z^\text{stu}_{t_0} \label{eq:stu-assign}\\ 
\mathcal{Q}^\text{teach}_{t_0} &= \arg\max_K M^\text{teach*} \cdot Z^\text{teach}_{t_0} \label{eq:teach-assign} 
\end{align}

where the optimal transport plans assigns each feature to its nearest cluster prototype. We guide the cluster assignments $\mathcal{Q}^{\text{teach}}_{t_0}$ using the motion trajectories $\text{Traj}_{t, i}$ obtained from eq.~\eqref{eq:traj}, where each tracked point retains its original cluster assignment as it moves along the trajectory:
\begin{equation}
\mathcal{Q}^{\text{teach}, i}_{t} = \mathcal{Q}^{\text{teach}, i}_{t_0}, \quad \forall (x_{i,t}, y_{i,t}) = \text{Traj}_{t,i} \label{eq:cotrack-clust-prop}
\end{equation}

\begin{figure}
\centering
\fig[1]{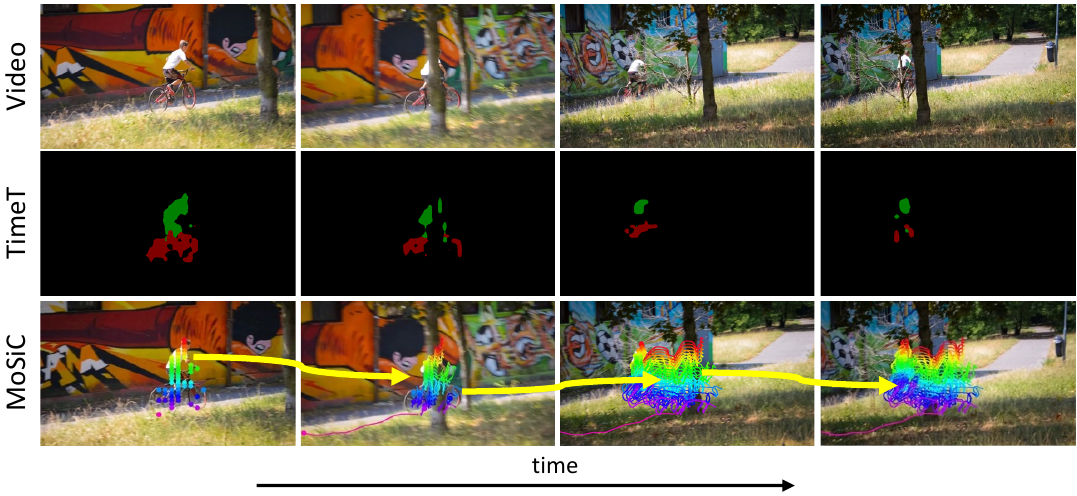}
\caption{ 
Previous methods such as TimeT lose precise object mapping after occlusions (e.g., a tree), rapid camera and object motion  -- resulting in degraded and coarse segmentation targets over long-range video clips. In contrast,~\methodname leverages stable point tracks, preserving consistency despite occlusions and motion via our trajectory based clustering loss eq.~\eqref{eq:final-loss}.
}
\label{fig:tracking}
\end{figure}

\subsection{Training Objective}

To ensure temporal consistency, we aim to align features of different views of the same object over time. Although motion trajectories provide the correspondence of object locations across frames, relying solely on trackers may not guarantee that features from different views are assigned to the same cluster. For each frame $t \in T$, let $\mathcal{Q}^{\text{stu},i}_{t_0}$ denote the cluster assignment for the $i $-th point in the initial frame $t_0$ from the student network, and $\mathcal{Q}^{\text{teach},i}_{t}$ denote the propagated cluster assignment for the $i $-th point in frame $t$ from the teacher network, as defined in eq~\eqref{eq:cotrack-clust-prop}.

We compute the cluster scores $S^{\text{stu},k, i}_{t_0}$ for each cluster $ k $ and point $ i $ in the frame $ t $ applying a softmax on the similarity between the $ Z^{\text{stu},i}_{t_0}$ and the cluster prototypes $ M^{\text{stu}*} $:

\begin{equation}
S^{\text{stu},k, i}_{t_0} = \frac{\exp(M^{\text{stu}*}\cdot  Z^{\text{stu},i}_{t_0})}{\sum_{k'} \exp(M^{\text{stu}*} \cdot  Z^{\text{stu},i}_{t_0})}
\end{equation}

The teacher's assignments are one-hot vectors, where $ \delta(\mathcal{Q}^{\text{teach},i}_{t} = k) $ equals 1 if the assigned cluster is $ k $ and 0 otherwise. The clustering loss is formulated as a cross-entropy loss, averaged over all visible trajectories and frames:

\begin{equation}
\mathcal{L}_{\text{clust}} (i) = - \sum_{t=1}^{T} \sum_{k=1}^{K}  v_{t,i} \cdot   \delta(\mathcal{Q}^{\text{teach},i}_{t} = k) \cdot \log(S^{\text{stu},k, i}_{t_0}) \label{eq:final-loss} 
\end{equation}

where $v$ is a visibility flag indicating whether the $ i $-th point is visible in frame $ t $. This ensures that the loss is computed only for points that are visible in each frame, thereby enhancing the robustness of our model to occlusions. Therefore, as shown in~\autoref{fig:tracking} we enforce that all patches along a trajectory, representing the same object or part as observed, are consistently clustered over time. This approach extends beyond individual images by maintaining consistency across pixel trajectories throughout the video sequence.

\begin{table*}[htb]
\centering
\caption{\textbf{In-context scene understanding benchmark.} We evaluate dense nearest neighbor retrieval performance across various training data proportions on ADE20K~\cite{zhou2017scene} and Pascal VOC~\cite{pascal-voc-2012}. Retrieved cluster maps are compared with the ground truth via Hungarian matching~\citep{kuhn1955hungarian}. We report mIoU; higher is better.}
\vspace{-6pt}
\label{tab:Hummingbird}
\footnotesize
{%
\begin{tabular}{lllcccccccc}
\toprule
& & & \multicolumn{4}{c}{\textsc{ADE20K}} & \multicolumn{4}{c}{\textsc{Pascal VOC}} \\
\cmidrule(r){4-7}\cmidrule(l){8-11}
\textsc{Method} & \textsc{Backbone} & \textsc{Params} & 1/128 & 1/64 & 1/8 & 1/1 & 1/128 & 1/64 & 1/8 & 1/1\\ 
\midrule
\addlinespace[1mm]
& \multicolumn{10}{c}{\textsc{Trained on Images}} \\
\addlinespace[1mm]
\quad DINO~\cite{caron2021emerging} & ViT-S/16 & 21M & 9.5 & 11.0 & 15.0 & 17.9 & 26.4 & 30.5 & 41.3 & 48.7 \\
\quad CrOC~\cite{stegmuller2023croc} & ViT-S/16 & 21M & 8.7 & 10.8 & 15.2 & 17.3 & 34.0 & 41.8 & 53.8 & 60.5 \\
\quad SelfPatch~\cite{yun2022patch} & ViT-S/16 & 21M & 10.0 & 10.9 & 14.7 & 17.7 & 28.4 & 32.6 & 43.2 & 50.8 \\
\quad Leopart~\cite{ziegler2022self} & ViT-S/16 & 21M & 12.9 & 14.8 & 19.6 & 23.9 & 44.6 & 49.7 & 58.4 & 64.5 \\
\quad CrlBo~\cite{lebailly2023cribo} & ViT-S/16 & 21M & 14.6 & 17.3 & 22.7 & 26.6 & 53.9 & 59.9 & 66.9 & 72.4 \\
\quad DINOv2~\cite{oquab2023dinov2} & ViT-S/14 & 21M & 22.8 & 26.4 & 33.5 & 38.8 & 56.0 & 62.4 & 72.3 & 77.0 \\
\addlinespace[1mm]
& \multicolumn{10}{c}{\textsc{Finetuned on Videos}} \\
\addlinespace[1mm]
\quad TimeT~\cite{salehi2023time} & ViT-S/16 & 21M & 12.1 & 14.1 & 18.9 & 23.2 & 38.1 & 43.8 & 55.2 & 62.3 \\
\rowcolor{lightcyan} \quad \methodname & ViT-S/14 & 21M & \textbf{23.8} & \textbf{27.4} & \textbf{35.7} & \textbf{40.7} & \textbf{62.5} & \textbf{66.6} & \textbf{74.7} & \textbf{78.2} \\
\midrule
\addlinespace[1mm]
& \multicolumn{10}{c}{\textsc{Trained on Images}} \\ 
\addlinespace[1mm]
\quad MAE~\cite{he2022masked} & ViT-B/16 & 85M & 10.0 & 11.3 & 15.4 & 18.6 & 3.5 & 4.1 & 5.6 & 7.0 \\
\quad DINO~\cite{caron2021emerging} & ViT-B/16 & 85M & 11.5 & 13.5 & 18.2 & 21.5 & 33.1 & 37.7 & 49.8 & 57.3 \\
\quad Leopart~\cite{ziegler2022self} & ViT-B/16 & 85M & 14.6 & 16.8 & 21.8 & 26.7 & 50.1 & 54.7 & 63.1 & 69.5 \\
\quad Hummingbird~\cite{balazevic2023towards} & ViT-B/16 & 85M & 11.7 & 15.1 & 22.3 & 29.6 & 50.5 & 57.2 & 64.3 & 71.8 \\
\quad CrlBo~\cite{lebailly2023cribo} & ViT-B/16 & 85M & 15.9 & 18.4 & 24.4 & 28.4 & 55.9 & 61.8 & 69.2 & 74.2 \\
\quad DINOv2~\cite{oquab2023dinov2} & ViT-B/14 & 85M & 24.2 & 27.6 & 34.7 & 39.9 & 55.7 & 61.8 & 72.4 & 77.1 \\
\addlinespace[1mm]
& \multicolumn{10}{c}{\textsc{Finetuned on Videos}} \\
\addlinespace[1mm]
\rowcolor{lightcyan} \quad \methodname & ViT-B/14 & 85M & \textbf{25.4} & \textbf{29.3} & \textbf{37.3} & \textbf{42.6} & \textbf{65.5} & \textbf{69.8} & \textbf{76.9} & \textbf{80.5} \\ 

\bottomrule
\end{tabular}
\vspace{-6pt}
}
\end{table*}

\section{Experiments}
\label{sec:exp}
\subsection{Experimental setup}
We follow TimeTuning~\cite{salehi2023time} and initialize~\methodname with a pretrained DINOv2~\cite{oquab2023dinov2} backbone and a frozen CoTracker-v3~\cite{karaev2024cotracker3} for point tracking. Training is conducted on YouTube-VOS~\cite{xu2018youtube}, one of the largest video segmentation datasets. For evaluation, we discard the projection head and use the teacher network as the feature extractor, following DINO~\cite{caron2021emerging}. Unless specified, mean intersection over union (mIoU) serves as the primary metric.

We assess linear segmentation on Pascal VOC~\cite{pascal-voc-2012}, COCO-Things~\cite{caesar2018coco}, and ADE20K~\cite{zhou2017scene} by training a segmentation head on frozen spatial features, and perform end-to-end segmentation using the Segmenter head~\cite{strudel2021segmenter}. For frozen clustering-based evaluations, we apply $K$-Means to spatial tokens, setting $K$ to both the number of ground-truth objects and larger values (300, 500), then match cluster maps to ground truth using Hungarian matching~\cite{kuhn1955hungarian}. Dense nearest-neighbor retrieval is used to evaluate on-context scene understanding~\cite{balazevic2023towards} for ADE20K and Pascal VOC. For unsupervised video semantic segmentation, we perform clustering and overclustering on YouTube-VOS and DAVIS~\cite{pont20172017}, aligning clusters with ground truth via Hungarian matching.

We evaluate \methodname across four dense benchmarks spanning images and videos, assessing unsupervised semantic segmentation across time and space, visual scene understanding, and feature transferability via linear segmentation. Additional experiments, including end-to-end finetuning results for object detection and segmentation, scaling to the large variant, image classification performance and pretraining with DINO, can be found in Appendix~\ref{appx:additional_experiments}

\begin{table*}[htb]   
\centering
\caption{\textbf{Unsupervised video semantic segmentation results} for clustering and over-clustering on DAVIS~\cite{pont20172017} and Youtube-VOS (YTVOS)~\cite{xu2018youtube}. 
For clustering, the Hungarian algorithm~\cite{kuhn1955hungarian} matches clusters (K) to ground truth (GT) per frame (F), clip (C), or dataset (D). 
For over-clustering: K=10 (F, C), K=200 (D, DAVIS), K=500 (D, YTVOS). All the models are ViT-S16, except for DINOv2 variants and \methodname, which are ViT-S14. We report mIoU and, unlike~\cite{salehi2023time}, do not use CBFE for any method to ensure consistency; higher is better.
\vspace{-6pt}
}
\footnotesize
\setlength{\tabcolsep}{5pt}
{
\begin{tabular}{l ccc c ccc c ccc c ccc}
\toprule
& \multicolumn{7}{c}{\textsc{Clustering}} & & \multicolumn{7}{c}{\textsc{Over-clustering}} \\
\cmidrule{2-8} \cmidrule{10-16}
& \multicolumn{3}{c}{\textsc{YTVOS}} & & \multicolumn{3}{c}{\textsc{DAVIS}} & & \multicolumn{3}{c}{\textsc{YTVOS}} & & \multicolumn{3}{c}{\textsc{DAVIS}} \\
\cmidrule{2-4} \cmidrule{6-8} \cmidrule{10-12} \cmidrule{14-16}
& \textit{F} & \textit{C} & \textit{D} & & \textit{F} & \textit{C} & \textit{D} & & \textit{F} & \textit{C} & \textit{D} & & \textit{F} & \textit{C} & \textit{D} \\
\midrule
\addlinespace[1mm]
\multicolumn{16}{c}{\textsc{Trained on Images}} \\
\addlinespace[1mm]
DINO~\cite{caron2021emerging} & 39.1 & 37.9 & 1.9 & & 30.2 & 31.0 & 1.6 & & 66.2 & 65.4 & 4.0 & & 56.9 & 54.9 & 17.9 \\
Leopart~\cite{ziegler2022self} & 39.2 & 37.9 & 11.7 & & 30.3 & 30.2 & 16.5 & & 64.5 & 62.8 & 15.5 & & 54.9 & 54.4 & 26.7 \\
DINOv2~\cite{oquab2023dinov2} & 56.3 & 55.5 & 12.8 & & 57.4 & 57.4 & 13.6 & & 62.8 & 62.5 & 17.3 & & 57.9 & 58.5 & 25.5 \\
DINOv2R~\cite{darcet2023vision} & 56.8 & 55.4 & 14.7 & & 57.3 & 57.8 & 14.3 & & 64.0 & 63.5 & 21.5 & & 58.1 & 59.5 & 27.2 \\
\midrule
\addlinespace[1mm]
\multicolumn{16}{c}{\textsc{Finetuned on Videos}} \\
\addlinespace[1mm] 
STEGO~\cite{hamilton2022unsupervised} & 41.5 & 40.3 & 2.0 & & 31.9 & 31.0 & 3.2 & & 58.1 & 54.3 & 5.1 & & 47.6 & 46.3 & 10.4 \\
DINO~\cite{caron2021emerging} & 37.2 & 36.1 & 1.2 & & 29.3 & 29.2 & 2.4 & & 53.1 & 50.9 & 1.3 & & 45.4 & 44.0 & 8.6 \\
Leopart~\cite{ziegler2022self} & 41.5 & 40.5 & 7.7 & & 37.5 & 36.5 & 12.6 & & 60.8 & 59.8 & 6.8 & & 53.7 & 53.1 & 16.8 \\
TimeT~\cite{salehi2023time} & 50.2 & 51.4 & 12.8 & & 49.5 & 49.2 & 12.8 & & 60.6 & 59.4 & 18.1 & & 55.9 & 57.6 & 22.0 \\
\rowcolor{lightcyan} \methodname & \textbf{60.6} & \textbf{59.6} & \textbf{18.4} & & \textbf{58.9} & \textbf{60.8} & \textbf{23.4} & & \textbf{66.8} & \textbf{65.6} & \textbf{24.4} & & \textbf{58.4} & \textbf{59.8} & \textbf{29.0} \\

\bottomrule
\vspace{-3pt}
\end{tabular}}
\label{table:clustering_video}
\end{table*}

\subsection{Unsupervised Video Object Segmentation}
\label{sec:exp-vos} To assess temporal coherence in~\methodname, we evaluate the clustering performance on DAVIS~\cite{pont20172017} and YouTube-VOS~\cite{xu2018youtube}. Following~\cite{salehi2023time}, clusters are aligned with ground-truth object masks using the Hungarian algorithm~\cite{kuhn1955hungarian}. We also evaluate performance under Over-Clustering, where a larger number of clusters is used. This is particularly relevant in dense self-supervised learning, as representations serve as feature descriptors for downstream tasks such as semantic segmentation and object detection.

From~\autoref{table:clustering_video}, \methodname-S14 outperforms TimeT~\cite{salehi2023time} by 8.7\% and 9.4\% mIoU on DAVIS and YouTube-VOS, respectively. Additionally, \methodname surpasses both DINOv2 and DINOv2-R by 4.3\% and 3.7\% on average across clustering and overclustering evaluations. These results highlight~\methodname’s ability to produce temporally consistent segmentations and generalize across diverse video datasets.

\vspace{-3pt}

\subsection{Visual In-Context learning}
\label{sec:exp-hb_eval}

We compare \methodname on the Hummingbird Benchmark~\cite{balazevic2023towards}, which evaluates in-context reasoning in vision models. Unlike traditional segmentation approaches, this task does not involve fine-tuning; instead, segmentation maps are generated by matching patch-level feature similarities between validation (query) and training (key) images. As shown in~\autoref{tab:Hummingbird}, \methodname consistently outperforms SOTA on Pascal-VOC~\cite{pascal-voc-2012} and ADE20K~\cite{zhou2017scene}, particularly in low-data settings. On Pascal VOC, \methodname-S14 outperforms DINOv2 by 6\% and DINOv2-R by nearly 10\% with 1/128th of the data and persists across all data regimes, with \methodname-S14 surpassing DINOv2 by 1\% even with full data. The larger variant, \methodname-B14, further improves performance, outperforming DINOv2-R by 10\% in the lowest-data setting and exceeding DINOv2 by 3\% with full data.

Similarly, on ADE20K, \methodname outperforms DINOv2 by 2\% across both variants. These results demonstrate \methodname’s ability to learn robust and transferable representations, especially in data-scarce scenarios where it significantly outperforms prior methods.

\vspace{-3pt}
\subsection{Frozen Clustering}
\label{sec:exp-frozen-cluster}
\begin{table*}[htb]
\centering
\caption{\textbf{Frozen clustering-based evaluations.} \textit{(a)} We evaluate the models using $K$-means with various clustering granularities $K$ on the features of Pascal VOC~\cite{pascal-voc-2012} and COCO-Things~\cite{caesar2018coco}. Cluster maps are matched to the  ground-truth via Hungarian matching~\cite{kuhn1955hungarian}, We report mIoU; higher is better. 
\textit{(b)} We post-process these maps for unsupervised semantic segmentation on Pascal VOC. All the models are ViT-S16, except for DINOv2 variants and \methodname, which are ViT-S14.
}
\vspace{-0.5em}
\begin{subtable}{0.65\textwidth}
\centering
\caption{Clustering}
\label{tab:clustering_label}
\footnotesize
{%
\begin{tabular}{lcccccc}
\toprule
\footnotesize
\setlength{\tabcolsep}{3pt}
& \multicolumn{3}{c}{\textsc{Pascal VOC}} & \multicolumn{3}{c}{\textsc{COCO-Things}} \\
\cmidrule(r){2-4} \cmidrule(l){5-7}
\textsc{Method}  & \textsc{K=100} & \textsc{K=300} &  \textsc{K=500} & \textsc{K=100} & \textsc{K=300} &  \textsc{K=500}\\ 
\midrule
\addlinespace[1mm]
& \multicolumn{6}{c}{\textsc{Trained on Images}} \\
\addlinespace[1mm]
\quad DINO~\cite{caron2021emerging}      & 10.1  & 13.9 & 17.3 & 14.4  & 18.8 & 19.2 \\
\quad CrOC~\cite{stegmuller2023croc}     & 10.2  & 16.4 & 20.0 & 22.4  & 14.7 & 18.1 \\
\quad iBOT~\cite{zhou2021ibot}           & 16.5  & 23.8 & 31.1 & 15.5  & 26.6 & 28.0 \\
\quad EVA-CLIP~\cite{sun2023eva}                  & 31.7 & 37.4 & 41.4 & 30.5 & 38.0 & 39.8   \\
\quad DINOv2R~\cite{darcet2023vision}           & 34.8 & 46.7 & 49.5 & 32.0 & 38.9 & 41.2 \\
\quad Leopart~\cite{ziegler2022self}     & 39.2 & 46.5 & 51.2 & 38.3 & 47.8 & 53.2 \\
\quad CrIBo~\cite{lebailly2023cribo}     & 40.3 & 51.3 & 54.5 & 40.2 & 46.0 & 48.3 \\
\quad DINOv2~\cite{oquab2023dinov2}             & 43.2 & 55.1 & 58.6 & 43.8 & 51.6 & 53.1 \\

\midrule
\addlinespace[1mm]
& \multicolumn{6}{c}{\textsc{Finetuned on Videos}} \\
\addlinespace[1mm]
\quad TimeT~\cite{salehi2023time}$^\dagger$       & 34.6 & 43.6 & 46.2 & 34.9 & 42.7 & 44.6 \\
\rowcolor{lightcyan}
\quad \methodname  & \textbf{50.0} & \textbf{58.8} & \textbf{60.2} & \textbf{45.8} & \textbf{53.2} & \textbf{54.9}\\
\bottomrule
\end{tabular}}
\end{subtable}
\hfill
\begin{subtable}{0.3\textwidth}
\centering
\caption{Semantic segmentation}
\label{tab:unsup_semantic_segmentation}
\footnotesize

\begin{tabular}{lc}
\toprule
\small
\textsc{Method} & \textsc{mIoU} \\
\midrule
\addlinespace[1mm]
\multicolumn{2}{c}{\textsc{Trained on Images}} \\
\addlinespace[1mm]
\quad MaskConstrast~\cite{van2021unsupervised}            & 35.1 \\
\quad DINOv2R~\cite{darcet2023vision}                  & 35.1 \\
\quad DINOv2~\cite{oquab2023dinov2}                   & 37.5 \\
\quad DeepSpectral~\cite{melas2022deep}            & 37.2 \\
\quad DINOSAUR~\cite{seitzer2022bridging}                 & 37.2 \\
\quad Leopart~\cite{ziegler2022self}                  & 41.7 \\
\quad COMUS~\cite{zadaianchuk2022unsupervised}                    & 50.0 \\
\midrule
\addlinespace[1mm]
\multicolumn{2}{c}{\textsc{Finetuned on Videos}} \\
\addlinespace[1mm]
\quad TimeT~\cite{salehi2023time} & {41.1} \\
\rowcolor{lightcyan}
\quad \methodname & \textbf{51.2}\\
\bottomrule
\end{tabular}
\end{subtable}
\end{table*}

We assess the quality of learned representations by clustering features and evaluating their ability to differentiate objects within datasets. Ideally, the patch features corresponding to the same object should be assigned to a single cluster. If representations capture finer details (e.g., hands or faces rather than entire persons), they should remain consistent across images. To analyze this, we extract dense features and apply $K$-Means clustering with varying $K$ values. Cluster maps are then aligned with ground truth using Hungarian matching~\cite{kuhn1955hungarian}. We evaluate performance when $K$ matches the number of ground-truth objects, as well as under Over-Clustering, similar to~\autoref{sec:exp-vos}.

From~~\autoref{tab:clustering_label} and~\autoref{tab:unsup_semantic_segmentation}, \methodname achieves the highest mIoU across all clustering granularities. For $K=300$, it outperforms DINOv2 by 3.7\% on Pascal VOC and 1.6\% on COCO-Things, indicating its ability to learn structured, discriminative representations. Under Over-Clustering ($K=500$), \methodname surpasses CrIBo by 5.7\% on Pascal VOC and 6.6\% on COCO-Things, demonstrating its capacity to encode fine-grained object structures. These results show \methodname’s effectiveness in distinguishing objects and their components, making it suitable for dense prediction tasks.

\subsection{Linear Segmentation}
\label{sec:exp-lin-seg}

\begin{table}[htb]
\centering
\caption{\textbf{Linear segmentation performance.} 
A linear segmentation head is trained on top of frozen spatial features. We report mIoU scores on 4 validation datasets.}
\vspace{-0.5em}
\label{tab:linear_seg}
\footnotesize
\resizebox{\columnwidth}{!}{%
\begin{tabular}{llcccc}
\toprule
\textsc{Method} & \textsc{Arch.} & \textsc{COCO-Things} & \textsc{COCO-Stuff} & \textsc{Pascal VOC} & \textsc{ADE20K} \\
\midrule

\multicolumn{6}{c}{\textsc{Trained on Images}} \\
\addlinespace[0.8mm]
DINO~\cite{caron2021emerging}  & ViT-S/16 & 43.9 & 45.9 & 50.2 & 17.5 \\
iBOT~\cite{zhou2021ibot}       & ViT-S/16 & 58.9 & 51.5 & 66.1 & 21.8 \\
CrOC~\cite{stegmuller2023croc} & ViT-S/16  & 64.3 & 51.2 & 67.4 & 23.1 \\
CrlBo~\cite{lebailly2023cribo} & ViT-S/16 & 64.3 & 49.1 & 71.6 & 22.7 \\
DINOv2~\cite{oquab2023dinov2}  & ViT-S/14 & 81.4 & 58.3 & 78.9 & 37.9 \\

\midrule

\multicolumn{6}{c}{\textsc{Finetuned on Videos}} \\
\addlinespace[0.8mm]
TimeT~\cite{salehi2023time} & ViT-S/16 & 58.2 & 48.7 & 66.3 & 20.7 \\
\rowcolor{lightcyan}
\methodname                & ViT-S/14 & \textbf{82.3} & \textbf{61.0} & \textbf{79.7} & \textbf{39.6} \\

\midrule

\multicolumn{6}{c}{\textsc{Trained on Images}} \\
\addlinespace[0.8mm]
DINO~\cite{caron2021emerging}  & ViT-B/16 & 55.8 & 51.2 & 62.7 & 23.6 \\
MAE~\cite{he2022masked}        & ViT-B/16 & 38.0 & 38.6 & 32.9 & 5.8 \\
iBOT~\cite{zhou2021ibot}       & ViT-B/16 & 69.4 & 55.9 & 73.1 & 30.1 \\
CrIBo~\cite{lebailly2023cribo} & ViT-B/16 & 69.6 & 53.0 & 73.9 & 25.7 \\
EVA-CLIP~\cite{sun2023eva}     & ViT-B/16 & 75.9 & 48.0 & 70.4 & 34.6 \\
DINOv2~\cite{oquab2023dinov2}  & ViT-B/14 & 84.0 & 58.9 & 80.3 & 42.6 \\

\midrule

\multicolumn{6}{c}{\textsc{Finetuned on Videos}} \\
\addlinespace[0.8mm]
\rowcolor{lightcyan}
\methodname           & ViT-B/14 & \textbf{85.8} & \textbf{61.4} & \textbf{81.5} & \textbf{43.6} \\

\bottomrule
\end{tabular}
}
\end{table}

We further evaluate \methodname for supervised semantic segmentation, keeping the pretrained backbone frozen while training a linear layer on top. Output features are upsampled via bilinear interpolation to match input dimensions, allowing for pixel-wise cross-entropy loss. Unlike fine-tuning, this setup provides a more reliable measure of the backbone’s learned representations. As shown in~\autoref{tab:linear_seg}, \methodname consistently outperforms SOTA segmentation models, including CrIBo, and surpasses DINOv2-R by up to 4\% on Pascal VOC and 3.3\% on ADE20K. Additionally, \methodname outperforms DINOv2 across all datasets, underscoring its ability to capture semantically meaningful features that enable strong segmentation performance with just a linear probe.

\subsection{Generalization to Diverse Vision Encoders}

We explore the generalization of \methodname to a diverse set of vision foundation models. We observe in~\autoref{table:generalization_backbones}, that \methodname enhances both in-context scene understanding and linear classification for vision models, as well as vision-language models such as EVA-CLIP~\cite{sun2023eva}, achieving improvements of 2\% and 3\% on average, respectively, using only video-based training. This demonstrates that videos can serve as a complementary source of information to further enrich image and language datasets.

\begin{table}[ht]
    \centering
    \caption{\textbf{Performance on In-context visual scene understanding (IC) and linear classification (LC) benchmarks with and without \methodname.} \methodname improves various baselines across segmentation datasets.}
    \resizebox{\columnwidth}{!}{%
    \begin{tabular}{lcccc}
        \toprule
        \multirow{2}{*}{\textsc{Method}} & \multicolumn{2}{c}{\textsc{IC}(1/1)} & \multicolumn{2}{c}{\textsc{LC}} \\
        \cmidrule(lr){2-3} \cmidrule(lr){4-5}
         & \textsc{ADE20K} & \textsc{Pascal VOC }& \textsc{ADE20K} & \textsc{Pascal VOC} \\
        \midrule
        \textsc{DINO-S16}          & 17.9 & 48.7 & 17.5 & 50.2 \\
        \rowcolor{lightcyan} + \methodname  & \textbf{24.9} & \textbf{64.5} & \textbf{22.3} & \textbf{67.3} \\
        \cmidrule{2-5}
        \textsc{EVA-CLIP-B14}~\cite{sun2023eva}    & 32.9 & 69.0 & 34.6 & 70.4 \\
        \rowcolor{lightcyan} + \methodname  & \textbf{35.0} & \textbf{70.0} & \textbf{40.1} & \textbf{73.4} \\
        \cmidrule{2-5}
        \textsc{DINOv2R-B14}       & 39.5 & 78.8 & 43.0 & 80.2 \\
        \rowcolor{lightcyan} + \methodname  & \textbf{43.7} & \textbf{79.3} & \textbf{44.4} & \textbf{81.1}  \\
        \bottomrule
    \end{tabular}
    }
    \label{table:generalization_backbones}
\end{table}

\begin{table*}[htb]
\caption{\textbf{Ablating the key parameters of \methodname} by training a linear layer on top of the frozen representations (Lin.) or using the in-context (IC) evaluation ~\cite{balazevic2023towards} on the validation images of Pascal VOC (PVOC) and ADE20K.}
\label{tab:full_abl}
\setlength{\tabcolsep}{3pt}
    \centering
    \begin{subtable}[t]{.32\textwidth}
        \centering
        \centering
\setlength{\tabcolsep}{6pt}
\caption{Mask Ratio}
\footnotesize
{\begin{tabular}{lcc}
    \toprule
    \textsc{Ratio} & \textsc{PVOC} & \textsc{ADE20K} \\
    \midrule
    No Mask   & 51.1 & 18.2 \\
    \rowcolor{lightcyan} 10\% & \textbf{51.5} & \textbf{18.6} \\
    20\% & 51.3 & 18.8 \\
    40\% & 49.9 & 18.5 \\
    \bottomrule
\end{tabular}
\label{table:mask_ratio}}

    \end{subtable}
    \begin{subtable}[t]{0.32\textwidth}
        \centering
        \centering
\setlength{\tabcolsep}{6pt}
\caption{EMA Teacher}
\footnotesize
{\begin{tabular}{lcc}
    \toprule
    \textsc{Teacher} & \textsc{PVOC} & \textsc{ADE20K} \\
    \midrule
    \ding{55} & 50.5 & 18.2 \\
\rowcolor{lightcyan}    \ding{51} &\textbf{51.5} & \textbf{18.6} \\
    \bottomrule
\end{tabular}
\label{table:ema_teacher}}

    \end{subtable}
    \begin{subtable}[t]{0.32\textwidth}
        \centering
        \centering
\setlength{\tabcolsep}{6pt}
\caption{Grid Size}
\footnotesize
{\begin{tabular}{lcc}
    \toprule
    \textsc{Grid} & \textsc{PVOC} & \textsc{ADE20K} \\
    \midrule
    $8\times8$   & 49.2 & 17.5\\
\rowcolor{lightcyan}    $16\times16$ & \textbf{51.5} & \textbf{18.6}\\
    $32\times32$ & 50.5 & 18.2\\
    \bottomrule
\end{tabular}
\label{table:grid_size}}

    \end{subtable}
    
    \par \vspace{3mm} %
    \begin{subtable}[t]{0.32\textwidth}
        \centering
        \centering
\setlength{\tabcolsep}{6pt}
\caption{Crop Scale}
\footnotesize
{\begin{tabular}{lcc}
    \toprule
    \textsc{Interval} & \textsc{PVOC} & \textsc{ADE20K} \\
    \midrule
    No crop  & 49.1 & 18.0 \\
    $[0.2, 1]$  & 50.8 & 18.5 \\
\rowcolor{lightcyan}    $[0.4, 1]$  & \textbf{51.5} & \textbf{18.6} \\
    $[0.6, 1]$  & 50.5 & 18.4 \\
    \bottomrule
\end{tabular}
\label{table:crop_scale}}

    \end{subtable}
    \begin{subtable}[t]{0.32\textwidth}
        \centering
        \centering
\setlength{\tabcolsep}{6pt}
\caption{Number of Prototypes}
\footnotesize
{\begin{tabular}{lcc}
    \toprule
    \textsc{Prototypes} & \textsc{PVOC} & \textsc{ADE20K} \\
    \midrule
    50  & 46.5 & 15.3\\
\rowcolor{lightcyan}    100  & \textbf{51.5} & \textbf{18.6} \\
    200  & 51.0 & 18.7 \\
    300  & 50.1 & 18.1 \\
    \bottomrule
\end{tabular}
\label{table:prototypes_abl}}

    \end{subtable}
    \begin{subtable}[t]{0.32\textwidth}
        \centering
        \centering
\setlength{\tabcolsep}{6pt}
\caption{Optimized points}
\footnotesize
{\begin{tabular}{lcc}
    \toprule
    \textsc{Batch} & \textsc{PVOC} & \textsc{ADE20K} \\
    \midrule
    All  & 50.5 & 18.0\\
\rowcolor{lightcyan}    Visible  & \textbf{51.5} & \textbf{18.6}\\
    \bottomrule
\end{tabular}
\label{table:optimized_points}}

    \end{subtable}
    \par \vspace{3mm} %
    \begin{subtable}[t]{0.32\textwidth}
        \centering
        
\centering
\caption{Clip length in second} 
\footnotesize
{\begin{tabular}{ccc}
\toprule
\multirow{1}{*}{$T$}  & \textsc{PVOC} & \textsc{ADE20K}\\
\midrule
\multirow{1}{*}{0.5} & 47.6 & 16.2\\
\multirow{1}{*}{1} & 50.5 & 18.2 \\
\multirow{1}{*}{1.6} & 51.0 & 18.2\\
\rowcolor{lightcyan}\multirow{1}{*}{3.2} & \textbf{51.5} & \textbf{18.6}\\
\bottomrule
\end{tabular}}
\label{table:clip_length}

    \end{subtable}
    \begin{subtable}[t]{0.32\textwidth}
        \centering
        \centering
\caption{Number of frames.}
\label{table:num_frame_effectiveness}
\footnotesize
    {\begin{tabular}{cccc}
    \toprule
    \textsc{\#frames} & \textsc{PVOC} & \textsc{ADE20K}\\
    \midrule
    2  & 39.3 & 13.1\\
    4  & 48.0 & 16.4\\
    8  & 50.4 & 18.2\\
\rowcolor{lightcyan}    12  & \textbf{51.5} & \textbf{18.6}\\
    \bottomrule
    \end{tabular}}
   \label{table:num_frames} 

    \end{subtable}
    \vspace{-6pt}
\end{table*}

\subsection{Ablations}
\label{sec:ablation}
We conduct ablations on key parameters of \methodname using the DINO~\cite{caron2021emerging} backbone trained on YouTube-VOS for 50 epochs with an image resolution of 224. We evaluate visual in-context learning on Pascal VOC12 and ADE20K, using a reduced memory size of 1/100 of the original setting in our primary experiments.
\vspace{-10pt}
\paragraph{Masking ratio.} We analyze the effect of different masking ratios in~\autoref{table:mask_ratio}. Moderate masking enhances semantic learning by preventing reliance on a limited set of patches. However, excessively high masking (e.g., 40\%) degrades performance by reducing available input information. We select a 10\% masking ratio due to its superior training stability over 20\%, despite both achieving similar performance.
\vspace{-3pt}

\paragraph{Crop scale.} \autoref{table:crop_scale} evaluates the impact of different cropping scales in augmentations. Moderate cropping yields the highest improvements, particularly due to the non-object-centric nature of YouTube-VOS, where aggressive cropping can make it difficult to extract meaningful regions consistently across frames. This differs from static-image datasets~\cite{ziegler2022self}.
\vspace{-3pt}
\paragraph{EMA teacher.} We assess the impact of an Exponential Moving Average (EMA) teacher in~\autoref{table:ema_teacher}. Incorporating a teacher network improves performance by 1\% on Pascal VOC and 0.4\% on ADE20K, aligning with findings from~\cite{caron2021emerging}, where EMA stabilization led to improved training consistency in self-supervised frameworks.

\vspace{-3pt}
\paragraph{Number of prototypes.} ~\autoref{table:prototypes_abl} shows that increasing the number of prototypes significantly enhances performance up to a threshold. The best results are achieved with 100 prototypes, beyond which improvements plateau. \methodname remains robust across a wide range (100–300 prototypes), indicating insensitivity to fine-tuning this parameter.
\vspace{-3pt}
\paragraph{Grid size.} \autoref{table:grid_size} shows the effect of different grid sizes for tracking initialized points. Denser grids improve performance, peaking at $16 \times 16$, after which performance slightly drops at $32 \times 32$. This decline is likely due to an incompatibility between patch-based feature extraction and point-wise tracking, where excessive density leads to misleading correspondences during training.
\vspace{-3pt}
\paragraph{Clip length.} We evaluate the effect of clip length in~\autoref{table:clip_length}, considering sequences of 12 frames. Longer clips improve performance by exposing the model to richer motion cues and diverse object transformations. We select a 3.2-second clip length as the optimal setting for all
\vspace{-3pt}
\paragraph{Number of clip frames.} \autoref{table:num_frames} explores the impact of clip frame count while keeping the time step per frame fixed at $3.2/12$ seconds. Increasing the frame count consistently improves performance, as it provides richer spatial and temporal signals during training. We use 12 frames as the optimal setting.
\vspace{-12pt}
\paragraph{Optimized points.} To improve tracking accuracy, we apply loss only to points that remain visible throughout the sequence. As shown in~\autoref{table:optimized_points}, this approach reduces false similarity enforcement, improving performance by up to 1%

\section{Conclusion}
\label{sec:conclusion}

In this paper, we introduced \methodname, a motion-based self-supervised learning approach that leverages dense point tracking to enforce temporal consistency in feature learning. By propagating cluster assignments along motion trajectories, our method effectively mitigates occlusions and long-range tracking drift, enhancing representation quality for dense prediction tasks. Extensive evaluations across unsupervised video-object segmentation, in-context learning, frozen clustering, and linear segmentation demonstrate that \methodname achieves state-of-the-art performance across both image and video benchmarks. Our results highlight the importance of motion cues in self-supervised representation learning and suggest that point tracking can serve as a powerful supervisory signal—enabling robust feature learning without requiring dense annotations.

\section{Acknowledgements}
Shashanka was supported by HPC resources from GENCI-IDRIS Grant 2024-AD011016023.

\clearpage

{
    \small
    \bibliographystyle{ieeenat_fullname}
    \bibliography{main}
}
\newpage
\clearpage
\newpage

\section{Appendix}

\subsection{Additional Experiments}
\label{appx:additional_experiments}

\begin{figure*}[htb]
    \centering
    \begin{center}
        \includegraphics[width=1\textwidth, trim={0cm 0cm 0cm 0cm},clip]{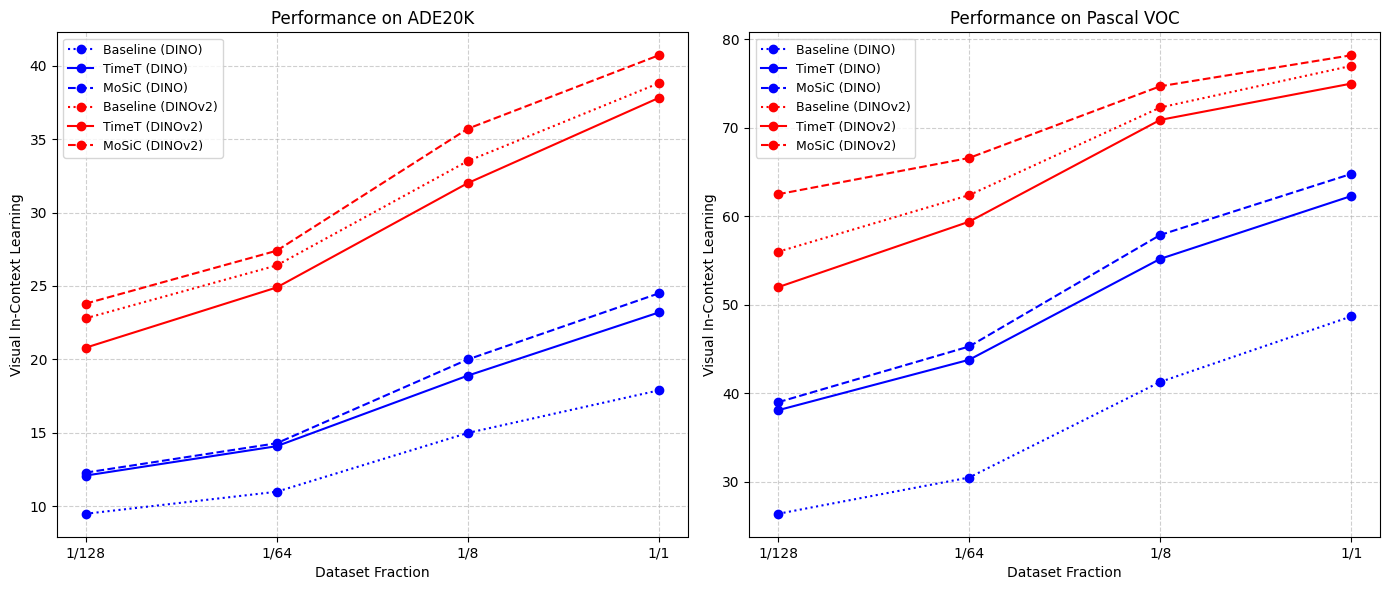}
    \end{center}
    \caption{\textbf{TimeT vs. \methodname.} As the figure shows, \methodname consistently improves both DINO and DINOv2 backbones, while TimeT only improves DINO.}
\label{fig:timet_vs_mosic}
\end{figure*}

\paragraph{Comparison to TimeT.} Here, we compare \methodname with TimeT, both initialized from the same DINO backbone. As shown in \autoref{fig:timet_vs_mosic}, \methodname consistently outperforms TimeT across Pascal VOC and ADE20K for both DINO and DINOv2 backbones. Notably, while TimeT struggles to further enhance strong vision backbones such as DINOv2 beyond the baseline, \methodname achieves consistent improvements across all benchmarks, demonstrating its greater generalizability.

\paragraph{Image classification performance.} As shown in \autoref{table:DINOv_vs_mosic_img_results}, while MoSiC significantly enhances the dense
understanding of DINOv2, it results in only a 0.8\% performance
reduction—negligible compared to the gains of up to 6\% in certain
benchmarks.

\begin{table}[h]
    \centering
    \caption{\textbf{\methodname vs. DINOv2 on downstream image classification.} \methodname greatly improves DINOv2's dense understanding while incurring a minimal 0.8\% performance reduction, outweighed by gains of up to 6\% in some benchmarks.}
    \resizebox{\columnwidth}{!}{%
    \begin{tabular}{lccccc}
        \toprule
        Method & Backbone & CIFAR-100 & ImgNet-100 & CIFAR-10 & Average \\
        \midrule
        DINOv2 & ViT-B/14 & 86.2\% & 90.5\% & 97.8\% & 91.5\% \\
        \rowcolor{lightcyan} \methodname & ViT-B/14 & 84.8\% & 89.9\% & 97.3\% & 90.7\% \\
        \bottomrule
    \end{tabular}
    }
    \label{table:DINOv_vs_mosic_img_results}
\end{table}

\paragraph{Full finetuning results.} In \autoref{table:mosic-obj-detect} and \autoref{tab:segmenter_finetuning}, we evaluate the performance of \methodname when used as the backbone for object detection and semantic segmentation in the full finetuning setting. For object detection ViT-Det~\cite{li2022exploring} and for semantic segmentation Segmenter~\cite{strudel2021segmenter} frameworks are used. As shown, our method consistently outperforms DINOv2 across all datasets and evaluation protocols, setting a new state-of-the-art—despite being fine-tuned solely on video data that differs in distribution from the evaluation datasets. Since full finetuning updates all model parameters, the superior performance of \methodname indicates that learning a more structured, object-aware feature space is achievable even without any image dataset, by finetuning on videos alone.

\begin{table}[h]
    \centering
    
    \caption{\textbf{\methodname vs. DINOv2 on object detection.} We use ViTDet~\cite{li2022exploring} on COCO~\cite{caesar2018coco} and report Average Precision (AP) on bounding-box object detection ($\text{AP}^\text{box}$) and instance segmentation ($\text{AP}^\text{mask}$).}
    \scriptsize
    \resizebox{\columnwidth}{!}{%
    \begin{tabular}{lcccc}
        \toprule
        Method & Backbone & Params & $\text{AP}^\text{box}$ & $\text{AP}^\text{mask}$ \\
        \midrule
        DINOv2      & ViT-S/14 & 21M & 42.5   & 36.7 \\
        \rowcolor{lightcyan}
        \methodname & ViT-S/14 & 21M   & 42.5   & 36.8   \\
        \midrule
        DINOv2      & ViT-B/14 & 85M   & 46.1   & 41.9 \\
        \rowcolor{lightcyan}
        \methodname & ViT-B/14 & 85M   & 46.4   & 42.0 \\
        \midrule
        DINOv2      & ViT-L/14 & 307M   & 51.6    & 45.9\\
        \rowcolor{lightcyan}
        \methodname & ViT-L/14 & 307M  & 51.8    & 46.0 \\
        \bottomrule
    \end{tabular}
    }
    \label{table:mosic-obj-detect}
\end{table}

\begin{table*}[t]
\centering
\caption{\textbf{Evaluation of full Finetuning with Segmenter.} Various backbones pre-trained with different self-supervised learning methods are fine-tuned using Segmenter~\cite{strudel2021segmenter}. The table shows the mIoU scores obtained on validation sets across 4 different datasets.}
\begin{tabular}{l l r c c c c}
\toprule
\textbf{Method} & \textbf{Backbone} & \textbf{Params} & \textbf{Pascal Context} & \textbf{Pascal VOC} & \textbf{COCO-Stuff} & \textbf{ADE20K} \\
\midrule
DINO      & ViT-S/16  & 21M & 46.0 & 80.3 & 43.2 & 43.3 \\
CrOC      & ViT-S/16  & 21M & 46.0 & 80.9 & 42.9 & 42.8 \\
TimeT     & ViT-S/16  & 21M & 47.4 & 80.4 & 43.1 & 43.5 \\
CrIBo     & ViT-S/16  & 21M & 49.3 & 82.3 & 43.9 & 45.2 \\
DINOV2   & ViT-S/14  & 21M & 58.0  & 80.4  & 42.1  & 44.0  \\
\rowcolor{lightcyan}
\methodname  & ViT-S/14  & 21M & \textbf{58.5}  & \textbf{81.1} & \textbf{42.5} & \textbf{44.4} \\
\midrule
DINO      & ViT-B/16     & 85M & 45.8 & 82.2 & 44.4 & 45.0 \\
MAE       & ViT-B/16     & 85M & 47.9 & 82.7 & 45.5 & 46.4 \\
CrIBo     & ViT-B/16     & 85M & 49.2 & 83.4 & 44.6 & 46.0 \\
DINOV2   & ViT-B/14      & 85M & 62.0 & 85.2  & 48.3  & 51.9  \\
\rowcolor{lightcyan}
\methodname  & ViT-B/14  & 85M & \textbf{62.5}  & \textbf{85.8}  & \textbf{48.6} & \textbf{52.4}      \\
\bottomrule
\end{tabular}
\label{tab:segmenter_finetuning}
\end{table*}

\paragraph{Generalization to large variants.} We report the performance for the large variant of \methodname for all the experiments in the paper, as shown by \autoref{table:mosic-obj-detect}, \autoref{tab:segmenter_finetuning}, \autoref{tab:Hummingbird_app}, \autoref{tab:clustering_label_app}, \autoref{tab:unsup_semantic_segmentation_app}, \autoref{table:clustering_video_app}, and \autoref{tab:linear_seg_app}. Our model can increase the performance of DINOv2-L even more than the small and base variants, showing the effectiveness of our method on large models. 

\begin{table*}[htb]
\centering
\caption{\textbf{In-context scene understanding benchmark.} We evaluate dense nearest neighbor retrieval performance across various training data proportions on ADE20K~\cite{zhou2017scene} and Pascal VOC~\cite{pascal-voc-2012}. Retrieved cluster maps are compared with the ground truth via Hungarian matching~\citep{kuhn1955hungarian}. We report mIoU; higher is better.}
\vspace{-6pt}
\label{tab:Hummingbird_app}
\footnotesize
{%
\begin{tabular}{lllcccccccc}
\toprule
& & & \multicolumn{4}{c}{\textsc{ADE20K}} & \multicolumn{4}{c}{\textsc{Pascal VOC}} \\
\cmidrule(r){4-7}\cmidrule(l){8-11}
\textsc{Method} & \textsc{Backbone} & \textsc{Params} & 1/128 & 1/64 & 1/8 & 1/1 & 1/128 & 1/64 & 1/8 & 1/1\\ 
\midrule
\addlinespace[1mm]
& \multicolumn{10}{c}{\textsc{Trained on Images}} \\
\addlinespace[1mm]
\quad DINO~\cite{caron2021emerging} & ViT-S/16 & 21M & 9.5 & 11.0 & 15.0 & 17.9 & 26.4 & 30.5 & 41.3 & 48.7 \\
\quad CrOC~\cite{stegmuller2023croc} & ViT-S/16 & 21M & 8.7 & 10.8 & 15.2 & 17.3 & 34.0 & 41.8 & 53.8 & 60.5 \\
\quad SelfPatch~\cite{yun2022patch} & ViT-S/16 & 21M & 10.0 & 10.9 & 14.7 & 17.7 & 28.4 & 32.6 & 43.2 & 50.8 \\
\quad Leopart~\cite{ziegler2022self} & ViT-S/16 & 21M & 12.9 & 14.8 & 19.6 & 23.9 & 44.6 & 49.7 & 58.4 & 64.5 \\
\quad CrlBo~\cite{lebailly2023cribo} & ViT-S/16 & 21M & 14.6 & 17.3 & 22.7 & 26.6 & 53.9 & 59.9 & 66.9 & 72.4 \\
\quad DINOv2~\cite{oquab2023dinov2} & ViT-S/14 & 21M & 22.8 & 26.4 & 33.5 & 38.8 & 56.0 & 62.4 & 72.3 & 77.0 \\
\addlinespace[1mm]
& \multicolumn{10}{c}{\textsc{Finetuned on Videos}} \\
\addlinespace[1mm]
\quad TimeT~\cite{salehi2023time} & ViT-S/16 & 21M & 12.1 & 14.1 & 18.9 & 23.2 & 38.1 & 43.8 & 55.2 & 62.3 \\
\rowcolor{lightcyan} \quad \methodname & ViT-S/14 & 21M & \textbf{23.8} & \textbf{27.4} & \textbf{35.7} & \textbf{40.7} & \textbf{62.5} & \textbf{66.6} & \textbf{74.7} & \textbf{78.2} \\
\midrule
\addlinespace[1mm]
& \multicolumn{10}{c}{\textsc{Trained on Images}} \\ 
\addlinespace[1mm]
\quad MAE~\cite{he2022masked} & ViT-B/16 & 85M & 10.0 & 11.3 & 15.4 & 18.6 & 3.5 & 4.1 & 5.6 & 7.0 \\
\quad DINO~\cite{caron2021emerging} & ViT-B/16 & 85M & 11.5 & 13.5 & 18.2 & 21.5 & 33.1 & 37.7 & 49.8 & 57.3 \\
\quad Leopart~\cite{ziegler2022self} & ViT-B/16 & 85M & 14.6 & 16.8 & 21.8 & 26.7 & 50.1 & 54.7 & 63.1 & 69.5 \\
\quad Hummingbird~\cite{balazevic2023towards} & ViT-B/16 & 85M & 11.7 & 15.1 & 22.3 & 29.6 & 50.5 & 57.2 & 64.3 & 71.8 \\
\quad CrlBo~\cite{lebailly2023cribo} & ViT-B/16 & 85M & 15.9 & 18.4 & 24.4 & 28.4 & 55.9 & 61.8 & 69.2 & 74.2 \\
\quad DINOv2~\cite{oquab2023dinov2} & ViT-B/14 & 85M & 24.2 & 27.6 & 34.7 & 39.9 & 55.7 & 61.8 & 72.4 & 77.1 \\
\addlinespace[1mm]
& \multicolumn{10}{c}{\textsc{Finetuned on Videos}} \\
\addlinespace[1mm]
\rowcolor{lightcyan} \quad \methodname & ViT-B/14 & 85M & \textbf{25.4} & \textbf{29.3} & \textbf{37.3} & \textbf{42.6} & \textbf{65.5} & \textbf{69.8} & \textbf{76.9} & \textbf{80.5} \\ 
\midrule
\addlinespace[1mm]
\quad DINOv2~\cite{oquab2023dinov2} & ViT-L/14 & 307M & 22.0 & 25.2 & 32.8 & 37.9 & 47.9  &  54.8 & 68.1 & 74.4 \\
\rowcolor{lightcyan} \quad \methodname & ViT-L/14 & 307M  & \textbf{24.7} & \textbf{28.1} & \textbf{35.7} & \textbf{41.0} &  \textbf{55.0} &  \textbf{62.0} & \textbf{73.6} & \textbf{78.5}\\

\bottomrule
\end{tabular}
\vspace{-6pt}
}
\end{table*}

\begin{table*}[htb]
\centering
\caption{\textbf{Frozen clustering-based evaluations.} \textit{(a)} We evaluate the models using $K$-means with various clustering granularities $K$ on the features of Pascal VOC~\cite{pascal-voc-2012} and COCO-Things~\cite{caesar2018coco}. Cluster maps are matched to the  ground-truth via Hungarian matching~\cite{kuhn1955hungarian}, We report mIoU; higher is better. 
\textit{(b)} We post-process these maps for unsupervised semantic segmentation on Pascal VOC. $^\dagger$: ViT-S/16.
}
\vspace{-0.5em}
\begin{subtable}{0.65\textwidth}
\centering
\caption{Clustering}
\label{tab:clustering_label_app}
\footnotesize
{%
\begin{tabular}{lcccccc}
\toprule
\footnotesize
\setlength{\tabcolsep}{3pt}
& \multicolumn{3}{c}{\textsc{Pascal VOC}} & \multicolumn{3}{c}{\textsc{COCO-Things}} \\
\cmidrule(r){2-4} \cmidrule(l){5-7}
\textsc{Method}  & \textsc{K=100} & \textsc{K=300} &  \textsc{K=500} & \textsc{K=100} & \textsc{K=300} &  \textsc{K=500}\\ 
\midrule
\addlinespace[1mm]
& \multicolumn{6}{c}{\textsc{Trained on Images}} \\
\addlinespace[1mm]
\quad DINO~\cite{caron2021emerging}$^\dagger$      & 10.1  & 13.9 & 17.3 & 14.4  & 18.8 & 19.2 \\
\quad CrOC~\cite{stegmuller2023croc}$^\dagger$     & 10.2  & 16.4 & 20.0 & 22.4  & 14.7 & 18.1 \\
\quad iBOT~\cite{zhou2021ibot}$^\dagger$           & 16.5  & 23.8 & 31.1 & 15.5  & 26.6 & 28.0 \\
\quad EVA-CLIP~\cite{sun2023eva}                   & 31.7 & 37.4 & 41.4 & 30.5 & 38.0 & 39.8   \\
\quad DINOv2R-S14~\cite{darcet2023vision}          & 34.8 & 46.7 & 49.5 & 32.0 & 38.9 & 41.2 \\
\quad Leopart~\cite{ziegler2022self}$^\dagger$     & 39.2 & 46.5 & 51.2 & 38.3 & 47.8 & 53.2 \\
\quad CrIBo~\cite{lebailly2023cribo}$^\dagger$     & 40.3 & 51.3 & 54.5 & 40.2 & 46.0 & 48.3 \\
\quad DINOv2-S14~\cite{oquab2023dinov2}            & 43.2 & 55.1 & 58.6 & 43.8 & 51.6 & 53.1 \\

\midrule
\addlinespace[1mm]
& \multicolumn{6}{c}{\textsc{Finetuned on Videos}} \\
\addlinespace[1mm]
\quad TimeT-S16~\cite{salehi2023time}$^\dagger$       & 34.6 & 43.6 & 46.2 & 34.9 & 42.7 & 44.6 \\
\rowcolor{lightcyan}
\quad \methodname-S14 & 50.0 & 58.8 & 60.2 & 45.8 & 53.2 & 54.9\\
\rowcolor{lightcyan}
\quad \methodname-B14   & 52.5   & 62.1  & 65.7     & 47.8   & 56.3   & 58.6   \\
\rowcolor{lightcyan}
\quad \methodname-L14    & \textbf{54.8}  & \textbf{69.6}  & \textbf{71.3}  & \textbf{51.1}  & \textbf{59.0}  & \textbf{60.8}  \\
\bottomrule
\end{tabular}}
\end{subtable}
\hfill
\begin{subtable}{0.3\textwidth}
\centering
\caption{Semantic segmentation}
\label{tab:unsup_semantic_segmentation_app}
\footnotesize

\begin{tabular}{lc}
\toprule
\small
\textsc{Method} & \textsc{mIoU} \\
\midrule
\addlinespace[1mm]
\multicolumn{2}{c}{\textsc{Trained on Images}} \\
\addlinespace[1mm]
\quad MaskConstrast~\cite{van2021unsupervised}$^\dagger$            & 35.1 \\
\quad DINOv2R-S14~\cite{darcet2023vision}                  & 35.1 \\
\quad DINOv2-S14~\cite{oquab2023dinov2}                  & 37.5 \\
\quad DINOv2-B14~\cite{oquab2023dinov2}                  & 36.7 \\
\quad DeepSpectral~\cite{melas2022deep}$^\dagger$             & 37.2 \\
\quad DINOSAUR$^\dagger$~\cite{seitzer2022bridging}                 & 37.2 \\
\quad Leopart~\cite{ziegler2022self}$^\dagger$                  & 41.7 \\
\quad COMUS~\cite{zadaianchuk2022unsupervised}$^\dagger$                    & 50.0 \\
\midrule
\addlinespace[1mm]
\multicolumn{2}{c}{\textsc{Finetuned on Videos}} \\
\addlinespace[1mm]
\quad TimeT~\cite{salehi2023time}$^\dagger$ & {41.1} \\
\rowcolor{lightcyan}
\quad \methodname-S14 & 51.2\\
\rowcolor{lightcyan}
\quad \methodname-B14 & 54.4\\
\rowcolor{lightcyan}
\quad \methodname-L14 & \textbf{57.1} \\
\bottomrule
\end{tabular}
\end{subtable}
\end{table*}

\begin{table*}[htb]   
\centering
\caption{\textbf{Unsupervised video semantic segmentation results} for clustering and over-clustering on DAVIS~\cite{pont20172017} and Youtube-VOS (YTVOS)~\cite{xu2018youtube}. 
For clustering, the Hungarian algorithm~\cite{kuhn1955hungarian} matches clusters (K) to ground truth (GT) per frame (F), clip (C), or dataset (D). 
For over-clustering: K=10 (F, C), K=200 (D, DAVIS), K=500 (D, YTVOS). We report mIoU; higher is better.
\vspace{-6pt}
}
\footnotesize
\setlength{\tabcolsep}{5pt}
{
\begin{tabular}{l ccc c ccc c ccc c ccc}
\toprule
& \multicolumn{7}{c}{\textsc{Clustering}} & & \multicolumn{7}{c}{\textsc{Over-clustering}} \\
\cmidrule{2-8} \cmidrule{10-16}
& \multicolumn{3}{c}{\textsc{YTVOS}} & & \multicolumn{3}{c}{\textsc{DAVIS}} & & \multicolumn{3}{c}{\textsc{YTVOS}} & & \multicolumn{3}{c}{\textsc{DAVIS}} \\
\cmidrule{2-4} \cmidrule{6-8} \cmidrule{10-12} \cmidrule{14-16}
& \textit{F} & \textit{C} & \textit{D} & & \textit{F} & \textit{C} & \textit{D} & & \textit{F} & \textit{C} & \textit{D} & & \textit{F} & \textit{C} & \textit{D} \\
\midrule
\addlinespace[1mm]
\multicolumn{16}{c}{\textsc{Trained on Images}} \\
\addlinespace[1mm]
DINO~\cite{caron2021emerging}$^\dagger$ & 39.1 & 37.9 & 1.9 & & 30.2 & 31.0 & 1.6 & & 66.2 & 65.4 & 4.0 & & 56.9 & 54.9 & 17.9 \\
Leopart~\cite{ziegler2022self}$^\dagger$ & 39.2 & 37.9 & 11.7 & & 30.3 & 30.2 & 16.5 & & 64.5 & 62.8 & 15.5 & & 54.9 & 54.4 & 26.7 \\
DINOv2-S14~\cite{oquab2023dinov2} & 56.3 & 55.5 & 12.8 & & 57.4 & 57.4 & 13.6 & & 62.8 & 62.5 & 17.3 & & 57.9 & 58.5 & 25.5 \\
DINOv2R-S14~\cite{darcet2023vision} & 56.8 & 55.4 & 14.7 & & 57.3 & 57.8 & 14.3 & & 64.0 & 63.5 & 21.5 & & 58.1 & 59.5 & 27.2 \\
\midrule
\addlinespace[1mm]
\multicolumn{16}{c}{\textsc{Finetuned on Videos}} \\
\addlinespace[1mm] 
STEGO~\cite{hamilton2022unsupervised}$^\dagger$ & 41.5 & 40.3 & 2.0 & & 31.9 & 31.0 & 3.2 & & 58.1 & 54.3 & 5.1 & & 47.6 & 46.3 & 10.4 \\
DINO~\cite{caron2021emerging}$^\dagger$ & 37.2 & 36.1 & 1.2 & & 29.3 & 29.2 & 2.4 & & 53.1 & 50.9 & 1.3 & & 45.4 & 44.0 & 8.6 \\
Leopart~\cite{ziegler2022self}$^\dagger$ & 41.5 & 40.5 & 7.7 & & 37.5 & 36.5 & 12.6 & & 60.8 & 59.8 & 6.8 & & 53.7 & 53.1 & 16.8 \\
TimeT~\cite{salehi2023time}$^\dagger$ & 50.2 & 51.4 & 12.8 & & 49.5 & 49.2 & 12.8 & & 60.6 & 59.4 & 18.1 & & 55.9 & 57.6 & 22.0 \\
\rowcolor{lightcyan} \methodname -S14 & \textbf{60.6} & \textbf{59.6} & 18.4 &  & 58.9 & \textbf{60.8} & 23.4 &  & 66.8 & 65.6 & 24.4 &  & \textbf{58.4} & 59.8 & 29.0 \\
\rowcolor{lightcyan} \methodname -B14 & 59.5 & 58.6 & 18.7 &  & 59.5 & 59.2 & 24.2 &  & \textbf{67.2} & 64.1 & 23.1 &  & 57.9 & 59.3 & 32.0 \\
\rowcolor{lightcyan} \methodname -L14 & 60.2 & 59.1 & \textbf{19.4} &  & \textbf{59.7} & 59.1 & \textbf{26.3} &  & 67.1 & \textbf{65.9} & \textbf{26.9} &  & 58.3 & \textbf{60.2} & \textbf{33.5} \\

\bottomrule
\vspace{-3pt}
\end{tabular}}
\label{table:clustering_video_app}
\end{table*}

\begin{table*}[htb]
\centering
\caption{\textbf{Linear segmentation performance.} 
A linear segmentation head is trained on top of the frozen spatial features obtained from 
different feature extractors. We report the mIoU scores achieved on the validation sets 
of 4 different datasets.}
\vspace{-0.5em}
\label{tab:linear_seg_app}
\footnotesize
\setlength{\tabcolsep}{3pt}
{%
\begin{tabular}{lllcccc}
\toprule
\textsc{Method} & \textsc{Backbone} & \textsc{Params} & 
\textsc{COCO-Things} & \textsc{COCO-Stuff} & \textsc{Pascal VOC} & \textsc{ADE20K} \\
\midrule

\addlinespace[1mm]
&\multicolumn{6}{c}{\textsc{Trained on Images}} \\
\addlinespace[1mm]
\quad DINO~\cite{caron2021emerging}
  & ViT-S/16 & 21M & 43.9 & 45.9 & 50.2 & 17.5 \\
\quad iBOT~\cite{zhou2021ibot}
  & ViT-S/16 & 21M & 58.9 & 51.5 & 66.1 & 21.8 \\
\quad CrOC~\cite{stegmuller2023croc}
  & ViT-S/16 & 21M & 64.3 & 51.2 & 67.4 & 23.1 \\
\quad CrlBo~\cite{lebailly2023cribo}
  & ViT-S/16 & 21M & 64.3 & 49.1 & 71.6 & 22.7 \\
\quad DINOv2~\cite{oquab2023dinov2}
  & ViT-S/14 & 21M & 81.4 & 58.3 & 78.9 & 37.9 \\
  
\midrule

\addlinespace[0.5mm]
&\multicolumn{6}{c}{\textsc{Finetuned on Videos}} \\
\addlinespace[0.5mm]

\quad TimeT~\cite{salehi2023time}
  & ViT-S/16 & 21M & 58.2 & 48.7 & 66.3 & 20.7 \\

\rowcolor{lightcyan}
\quad \methodname & ViT-S/14 & 21M & \textbf{82.3} & \textbf{61.0} &  \textbf{79.7} & \textbf{39.6} \\

\midrule

\addlinespace[1mm]
&\multicolumn{6}{c}{\textsc{Trained on Images}} \\
\addlinespace[1mm]
\quad DINO~\cite{caron2021emerging}
  & ViT-B/16 & 85M & 55.8 & 51.2 & 62.7 & 23.6 \\
\quad MAE~\cite{he2022masked}
  & ViT-B/16 & 85M & 38.0 & 38.6 & 32.9 & 5.8 \\
\quad iBOT~\cite{zhou2021ibot}
  & ViT-B/16 & 85M & 69.4 & 55.9 & 73.1 & 30.1 \\
\quad CrIBo~\cite{lebailly2023cribo}
  & ViT-B/16 & 85M & 69.6 & 53.0 & 73.9 & 25.7 \\
\quad EVA-CLIP~\cite{sun2023eva}
  & ViT-B/14 & 86M  & 75.9  & 48.0  & 70.4   & 34.6  \\
\quad DINOv2~\cite{oquab2023dinov2}
  & ViT-B/14 & 85M & 84.0 & 58.9 & 80.3 & 42.6 \\
\midrule
\addlinespace[0.5mm]
&\multicolumn{6}{c}{\textsc{Finetuned on Videos}} \\
\addlinespace[0.5mm]
\rowcolor{lightcyan}
\quad \methodname & ViT-B/14    & 85M & \textbf{85.8} & \textbf{61.4} & \textbf{81.5} & \textbf{43.6} \\
\midrule
\quad DINOv2~\cite{oquab2023dinov2} & ViT-L/14 & 307M & 83.8  & 58.0  & 79.7  & 41.8  \\
\rowcolor{lightcyan}
\quad \methodname & ViT-L/14 & 307M  & \textbf{85.7}  & \textbf{61.5}  & \textbf{81.8}  & \textbf{44.7}   \\

\bottomrule
\end{tabular}
}
\end{table*}

\paragraph{Tracker ablation.} We train and evaluate~\methodname using both RAFT and CoTrackerv2 to investigate the effect of different trackers on our method. From~\autoref{tab:tracker_abl}, \methodname improves over DINOv2 with both these point trackers. \methodname is compatible with multiple point trackers, yet the better the tracker, the better \methodname performs.

\begin{table}[h]
\centering
\footnotesize
\setlength{\tabcolsep}{6pt}
\caption{\textbf{Choice of Point Trackers.} In-context scene understanding results (mIoU) on PVOC and ADE20K. Higher is better.}
\begin{tabular}{lcc}
    \toprule
    \textbf{Method} & \textbf{PVOC (mIoU)} & \textbf{ADE20K (mIoU)} \\
    \midrule
    DINOv2         & 77.0 & 38.8 \\
    \midrule
    RAFT           & 77.6 & 39.7 \\
    CoTracker2     & 78.0 & 40.2 \\
\rowcolor{lightcyan}    CoTracker3     & \textbf{78.2} & \textbf{40.7} \\
    \bottomrule
\end{tabular}
\label{tab:tracker_abl}
\end{table}

\subsection{Additional Visualizations}

\paragraph{Hummingbird qualitative results for \methodname.} Here, we show the qualitative results of our method for the visual in-context learning evaluation (Hummingbird benchmark) shown by \autoref{tab:Hummingbird}. As shown, although \methodname is finetuned on YTVOS videos—which differ significantly in distribution from Pascal—it still produces accurate and precise semantic segmentation maps, characterized by distinct IDs and tight boundaries.

\begin{figure*}[t]
  \centering
  \includegraphics[width=\textwidth, trim=3.5cm 0.2cm 10cm 0cm,clip]{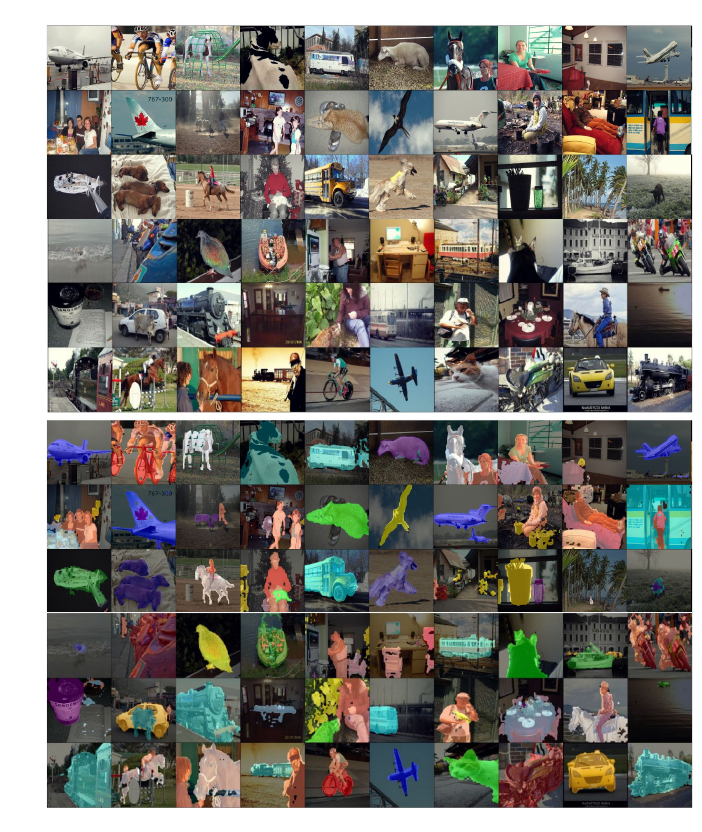}
  \caption{\textbf{Hummingbird qualitative results for \methodname on Pascal VOC.} The first group of rows displays the images, the second shows the corresponding per-image masks, and the third overlays the masks on the images, aligned by their semantic IDs. As shown, although \methodname is finetuned on YTVOS videos—which differ significantly in distribution from Pascal—it still produces accurate and precise semantic segmentation maps, characterized by distinct IDs and tight boundaries.}
  \label{fig:hb_vis}
\end{figure*}

\paragraph{Overclustering visualizations for \methodname.}

\begin{figure*}[!htb]
    \centering
    \includegraphics[width=\textwidth, trim=0cm 0cm 0cm 0cm, clip]{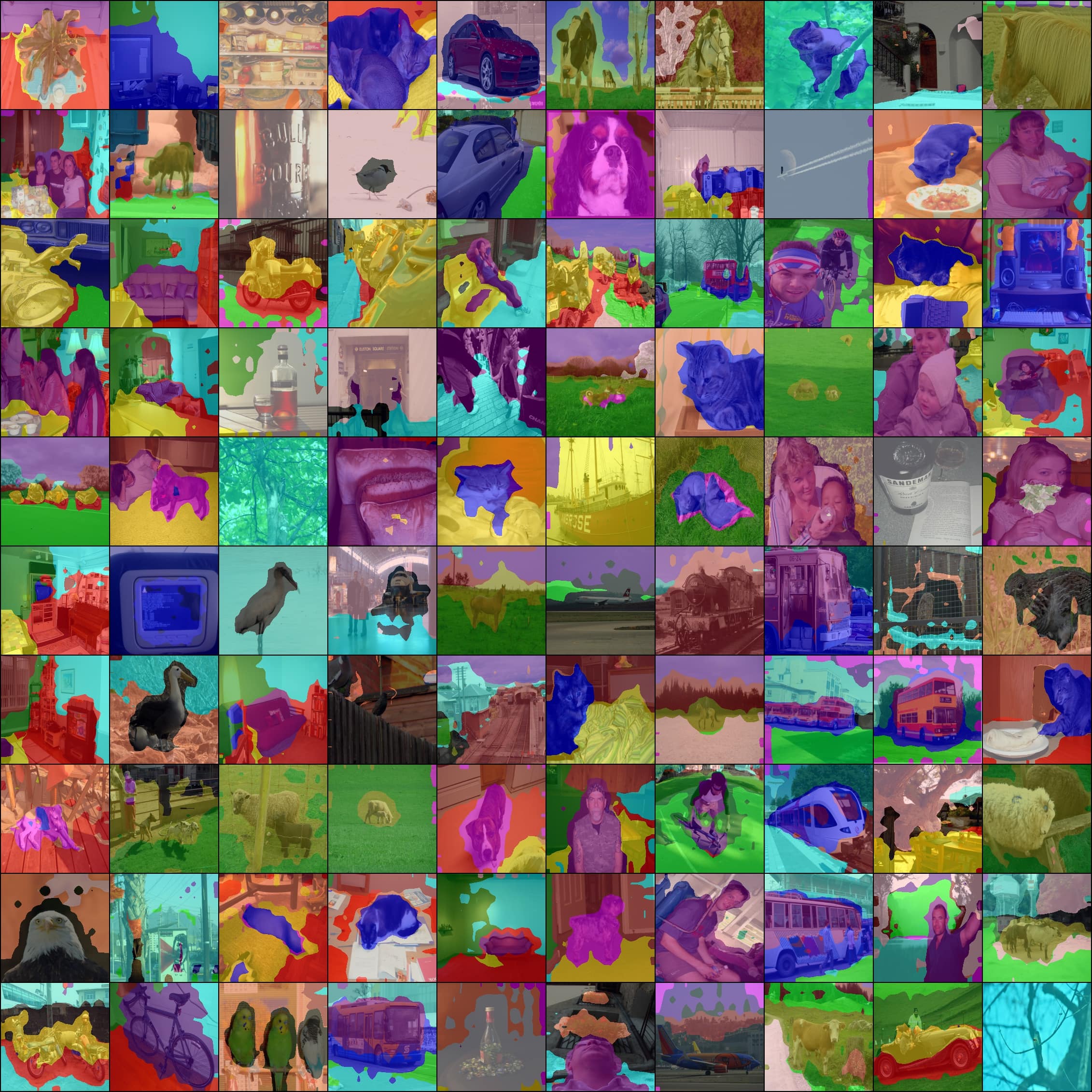}
    \caption{\textbf{{\methodname} overclustering visualizations on Pascal for \(\mathbf{K}=50\).} {\methodname} not only localizes objects precisely but also identify them with different cluster ids demonstrated by different colors. For instance, classes such as birds, motorcycles, dogs, cats, and cars are clearly identifiable. }
    \label{fig:overclustering}
\end{figure*}

We present the overclustering qualitative results of \methodname on Pascal VOC for $K=50$, approximately twice the number of object categories in the dataset, as shown in \autoref{fig:overclustering}. As shown, \methodname not only localizes objects precisely but also identify them with different cluster ids demonstrated by different colors. For instance, classes such as birds, motorcycles, dogs, cats, and cars are clearly identifiable.

\begin{figure}[h]
    \centering
    \includegraphics[width=1\linewidth]{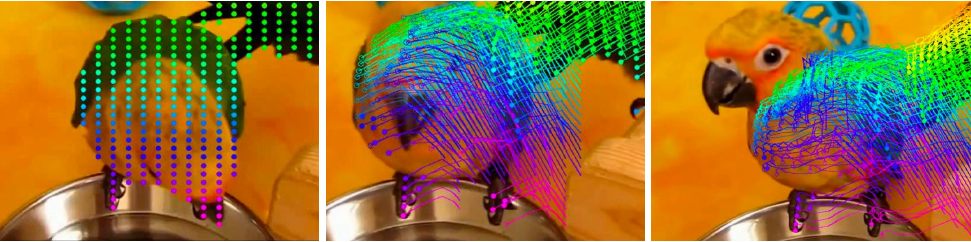}
    \caption{\textbf{Incorrect trajectories. }The point tracks that start from the beak of the bird and eventually track the bowl. This happens due to the ambigious motion of the bird's head.} 
    \label{fig:failure}
\end{figure}

\paragraph{Failure cases.} While~\methodname is robust to occlusions, it can struggle when motion cues are subtle or ambiguous, reducing the effectiveness of clustering supervision. In rare cases of tracker failure—due to extreme motion, prolonged occlusions, or appearance changes—noisy trajectories may propagate incorrect signals (\autoref{fig:failure}). Overly dense tracking (e.g., $32 {\times} 32$ grids) can similarly introduce spurious correspondences (Ablations - Table 5c). These issues arise from dependence on accurate tracking and clustering, though~\methodname mitigates them by focusing losses on visible points. 

\subsection{Dense Post-Pretraining \label{app:detailstraining}}

\paragraph{Implementation Framework}
We implement our model in Python using Torch \cite{paszke2019pytorch}.

\paragraph{Datasets}
Our pretraining datasets include COCO \cite{caesar2018coco} and the ImageNet-100 subset of the original ImageNet \cite{russakovsky2015imagenet}. COCO comprises approximately 118,000 scene-centric images, while ImageNet-100 contains around 100,000 object-centric images.

\paragraph{Network Architecture}
We use vision transformers as our backbone, specifically training on ViT-Small and ViT-Base \cite{dosovitskiy2020image}. Following \cite{caron2021emerging,grill2020bootstrap}, we adopt a student-teacher setup, where the teacher's weights are updated via the exponential moving average of the student's weights.

\paragraph{Projection Head}
As in \cite{caron2021emerging}, our projection head comprises three linear layers with a hidden dimensionality of 2048, Gaussian error linear units (GELU) as the activation function \cite{hendrycks2016gaussian}, and an output dimensionality of 256.

\paragraph{Optimization}  
We train both network sizes using a cosine learning rate schedule that decays to $0$ over $8$ epochs for DINOv2 and $100$ epochs for DINO, except for the ablation studies, where we use $50$ epochs on DINO. The initial learning rate for the projection head is set to $1\mathrm{e}{-4}$ across all experiments, while the backbone’s learning rate is $1\mathrm{e}{-5}$. The teacher’s weights are updated using an exponential moving average with the coefficient $0.99$. We optimize our model with Adam \cite{kingma2014adam}, applying a cosine weight decay schedule.

\subsection{Evaluation Setup \label{appendix_a:evaluation_setup}}

\paragraph{Visual In-Context Learning}  
The Dense Nearest Neighbor Retrieval Evaluation is a retrieval-based scene understanding benchmark introduced by \cite{balazevic2023towards}. It aims to assess the scene understanding capabilities of a dense image encoder. The evaluation follows these steps:

\begin{enumerate}
    \item \textbf{Memory Bank Construction}: Given a dataset of images with dense annotations, two memory banks are created. One stores image patch features extracted from the spatial output of a dense encoder applied to the training images, while the other stores the corresponding patch labels from the dataset annotations.

    \item \textbf{Query Processing}: For each image in the validation set, the spatial output of the dense image encoder is computed. For each patch representation, the $k$ nearest neighbors are retrieved from the feature memory bank. The labels of these nearest neighbors are then aggregated to construct the query’s annotation.

    \item \textbf{Comparison}: The generated annotation for the image is compared against the ground truth annotation to evaluate performance.
\end{enumerate}

Since the original implementation by \cite{balazevic2023towards} is unavailable, we use the open-source implementation from \cite{pariza2024hbird}. This implementation adheres to the original authors’ methodology, including the use of the ScaNN library \cite{guo2020avq} for efficient nearest neighbor retrieval. For our experiments, we follow the setup used by the Hummingbird Model authors \cite{balazevic2023towards}, utilizing a memory size of $10,240,000$ and configuring ScaNN with $30$ nearest neighbors, consistent with the Hummingbird model evaluation.

Final results are reported as mean Intersection over Union (mIoU) on four different fractions of two datasets: Pascal VOC 2012 \cite{pascal-voc-2012} and ADE20K \cite{zhou2017scene}. The sub-sampling factors considered are $1$, $8$, $64$, and $128$. For factors greater than $1$, results are averaged over five different seeds. These dataset subsets are created by randomly and uniformly selecting a distinct set of images from the training split, ensuring an approximately equal number of unique images per annotation label. For instance, for the $1/128$ fraction of Pascal VOC 2012, we sample around $83$ images, ensuring that each of the $20$ labels (excluding the background) appears in at least $4$ different images within the subset.

\paragraph{Overclustering}  
Following \cite{ziegler2022self}, we conduct the Overclustering experiment by applying $K$-Means clustering (using FAISS \cite{johnson2019billion}) to all spatial tokens from our backbone, omitting the projection head. The resulting clusters are then mapped to the dataset’s ground-truth classes using a two-step process: first, greedy matching based on pixel-level precision, followed by Hungarian matching \cite{kuhn1955hungarian} on the combined cluster maps. This procedure ensures that the evaluation metric remains permutation-invariant \cite{ji2019invariant}.  

Input images are resized to $448 \times 448$, while overclustering is performed on downsampled $100 \times 100$ masks to accelerate Hungarian matching. The final results are reported as the average mean Intersection over Union (mIoU) over five different seeds across four datasets: COCO-Thing and COCO-Stuff \cite{caesar2018coco}, Pascal VOC 2012 \cite{pascal-voc-2012}, and ADE20K \cite{zhou2017scene}.

\paragraph{Linear Segmentation}  
For linear segmentation, we follow the setup from Leopart \cite{ziegler2022self}. Specifically, we process $448 \times 448$ images through our backbone to extract spatial feature outputs, apply bilinear interpolation to align them with the mask resolution, and use a linear head to generate segmentation predictions. These predictions are then compared to the ground-truth segmentation masks and optimized using cross-entropy loss.  

To speed up training, we downsample the segmentation masks to $100 \times 100$. The linear head is optimized using Stochastic Gradient Descent with a weight decay of $0.0001$, a momentum of $0.9$, and a step-based learning rate scheduler. We find that a learning rate of $0.01$ works well across the backbone models we evaluate. The linear heads are fine-tuned for $20$ epochs.  

We train and evaluate the linear heads on four datasets: Pascal VOC 2012 \cite{pascal-voc-2012}, subsets of COCO-Thing and COCO-Stuff , and ADE20K \cite{zhou2017scene}.

\paragraph{Fully Unsupervised Semantic Segmentation}  

To further assess the scene understanding capabilities of our method, we evaluate it using the Fully Unsupervised Semantic Segmentation Evaluation \cite{ziegler2022self}. This evaluation consists of two components: Cluster-based Foreground Extraction (CBFE) and Overclustering with Community Detection (CD).  

CBFE clusters the spatial outputs of a model over a dataset and assigns each cluster as either background (\textit{bg}) or foreground (\textit{fg}). The foreground-background separation is guided by attention maps from a Vision Transformer, which provide cues for distinguishing between fg/bg regions. To construct the final hard fg-bg assignment, we average the attention heads, apply Gaussian filtering with a $7 \times 7$ kernel, and retain $70\%$ of the attention mass to generate the binary mask. The rest of the configurations follow the original setup \cite{ziegler2022self}.  

The CD metric \cite{ziegler2022self} leverages local co-occurrence statistics among clusters to identify and categorize objects. This approach is entirely label-free, relying only on the local co-occurrence of clusters within an image based on an information-theoretic definition of network communities. Our CD evaluation configurations remain consistent with those used in Leopart \cite{ziegler2022self}.  

We use the implementation from Leopart \cite{ziegler2022self} and apply CBFE and CD to the non-augmented (\textit{train}) split of Pascal VOC 2012 \cite{pascal-voc-2012}, evaluating on its full validation set. For CD, we report the best results across 10 seeds obtained via a hyperparameter search. Our best-performing parameters for CD+CBFE are: \textit{weight\_threshold} = $0.07$, \textit{markov\_time} = $1.6$, and \textit{k\_community} = $169$.

\section{Dataset Details}
\label{appendix:datasets}

\subsection{Image Datasets}

\begin{figure*}[t]
  \centering
  \includegraphics[width=\textwidth, trim=0cm 0cm 0cm 0cm,clip]{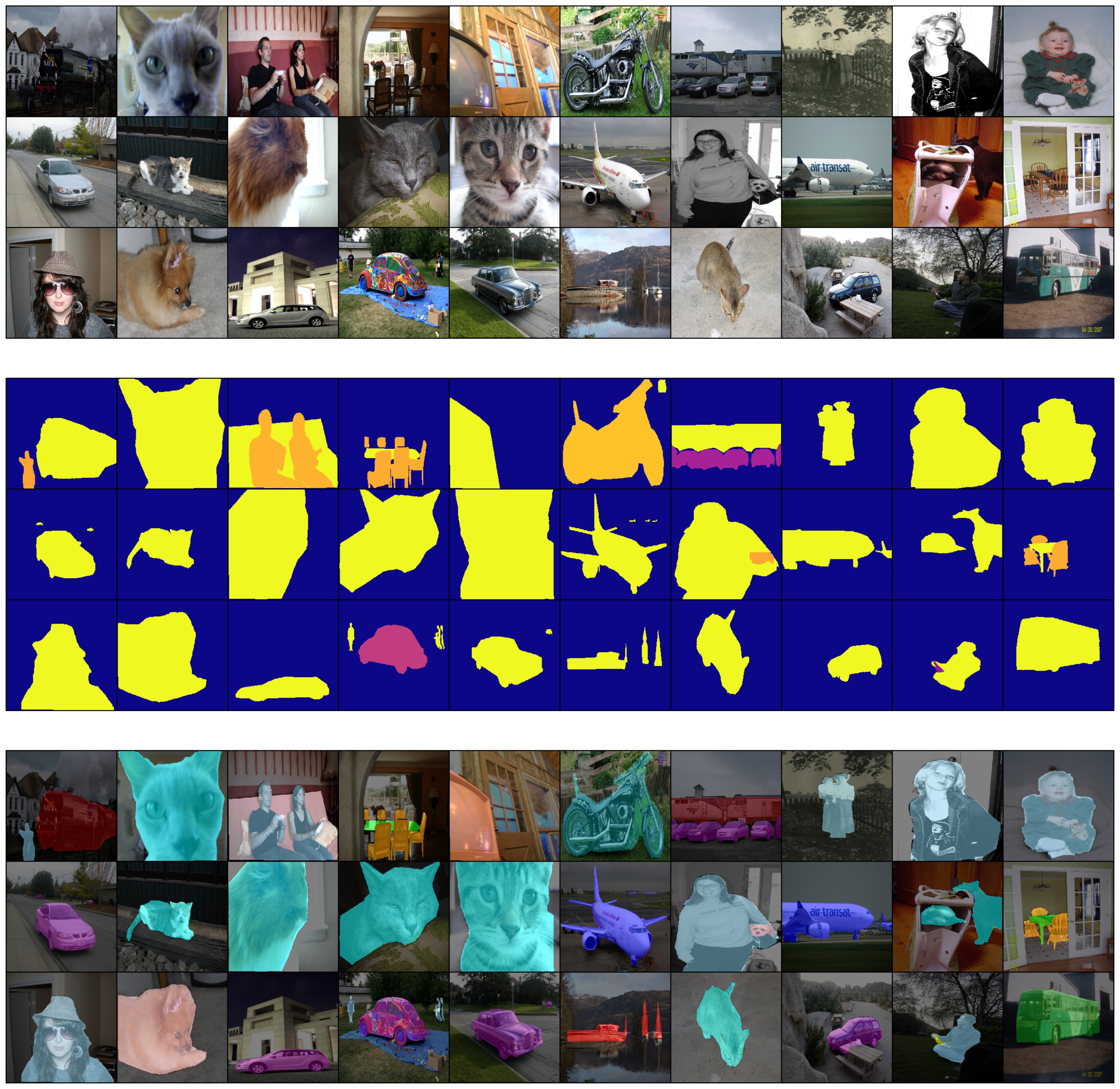}
  \caption{\textbf{Pascal VOC visualizations.} The first group of rows displays the images, the second shows the corresponding per-image masks, and the third overlays the masks on the images, aligned by their semantic IDs. These images and their ground truth segmentation maps are used for our tasks, such as visual in-context learning and linear segmentation.}
  \label{fig:pvoc_mem}
\end{figure*}

\noindent \textbf{Pascal VOC 2012} \cite{pascal-voc-2012}  
This dataset, using the latest trainaug split, consists of 10,582 images with annotations spanning 21 classes, including one background class. The validation set contains 1,449 images. Following \cite{van2021unsupervised}, we ignore unlabeled objects as well as the boundary class. For hyperparameter tuning of the fully unsupervised segmentation method \cite{ziegler2022self} applied to our method, we use the PVOC12 train split with 1,464 images. \autoref{fig:pvoc_mem} provides an overview of dataset images overlaid with annotations.

\noindent \textbf{COCO-Stuff 164K} \cite{caesar2018coco}  
COCO-Stuff is a scene-understanding dataset with labels across 91 "stuff" categories and 80 "thing" categories. The training set contains 118,000 images, while the validation set includes 5,000 images. Following \cite{ziegler2022self}, we use the COCO benchmark in two ways to better isolate object definitions.  

First, we extract stuff annotations, which represent objects without clear boundaries and are often part of the background, using COCO-Stuff annotations \cite{caesar2018coco}. We then consolidate the 91 detailed labels into 15 broader categories as described in \cite{ji2019invariant}, assigning the general label “other” to non-stuff objects since this label lacks specific semantic meaning. Non-stuff objects are ignored during training and evaluation. In our work, we refer to this dataset version as COCO-Stuff, which is used for Overclustering and Linear Segmentation (see Appendix \ref{appendix_a:evaluation_setup}).  

Next, we extract foreground annotations using the panoptic labels from \cite{kirillov2019panoptic}. Instance-level annotations are merged into object categories using the authors' provided script. Additionally, we consolidate the 80 detailed categories into 12 broad object classes. The background class is ignored during training and evaluation. This results in the COCO-Thing dataset version, which we use for Overclustering and Linear Segmentation (see Appendix \ref{appendix_a:evaluation_setup}).  

\noindent \textbf{ADE20K} \cite{zhou2017scene}  
ADE20K is a diverse dataset for semantic segmentation, containing finely detailed labels across 150 unique semantic categories. These include "stuff" labels such as sky and grass, as well as distinguishable objects like people and cars. The dataset features a wide range of scenes, with 20,210 images in the training set and 2,000 images in the validation set, making it one of the most challenging and diverse benchmarks for scene understanding. We use the full dataset in our experiments while ignoring the \textit{others} label.  

\begin{figure}[htb]
    \centering
    \begin{center}
        \includegraphics[width=1\linewidth, trim={0.2cm 0.2cm 0.2cm 0.2cm},clip]{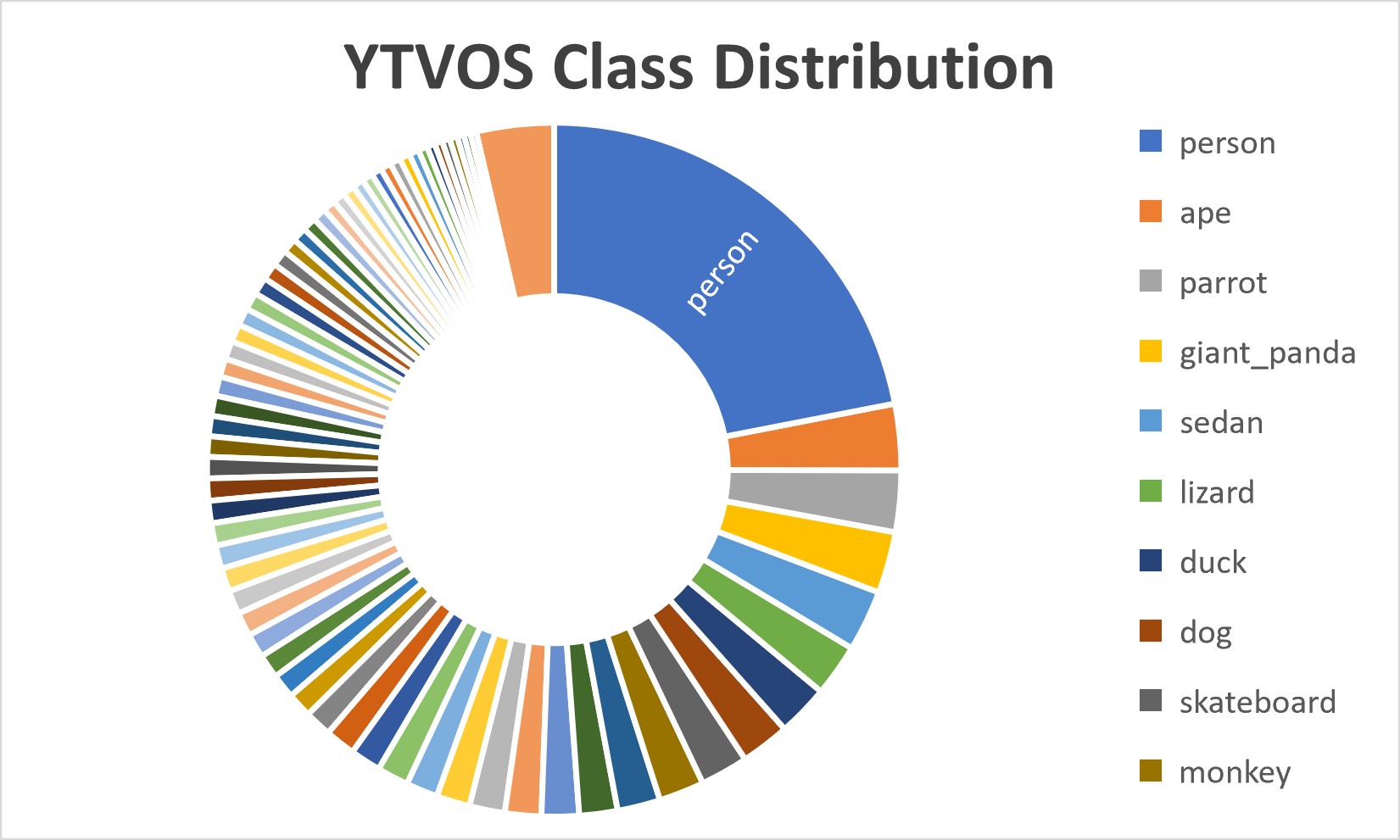}
    \end{center}
    \caption{The distribution of classes in YouTube-VOS. Some of the more dominant classes are labeled. The image is taken from~\cite{salehi2023time}}
\label{fig:appendix_class_distribution}
\end{figure}

\subsection{Video Datasets}

\begin{figure*}[t]
  \centering
  \includegraphics[width=\textwidth, trim=0cm 0cm 0cm 0cm,clip]{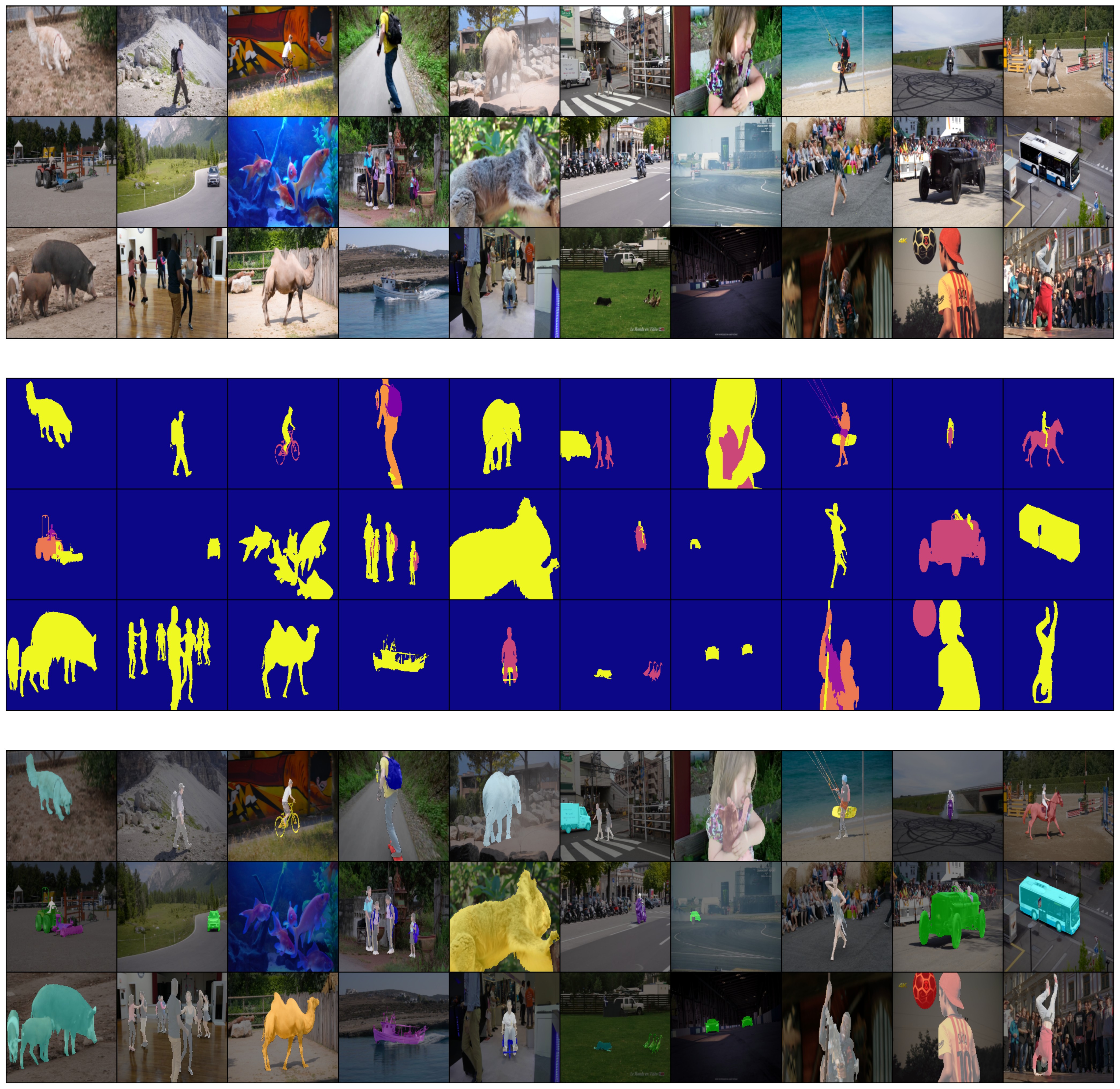}
  \caption{\textbf{DAVIS visualizations.} The first group of rows displays the images, the second shows the corresponding per-image masks, and the third overlays the masks on the images, aligned by their semantic IDs. These images and their ground truth segmentation maps are used for our tasks.}
  \label{fig:davis_mem}
\end{figure*}

\noindent \textbf{\textit{DAVIS17}}~\cite{pont20172017} is a benchmark for video object segmentation, consisting of 150 videos, with 60 allocated for training, 30 for validation, and 60 for testing. Since ground-truth foreground masks are only available for the first frames of the test set, the validation set is used for evaluation.  An overview of dataset frames and annotations can be viewed in \autoref{fig:davis_mem}.

\begin{figure*}[t]
  \centering
  \includegraphics[width=\textwidth, trim=0cm 0cm 0cm 0cm,clip]{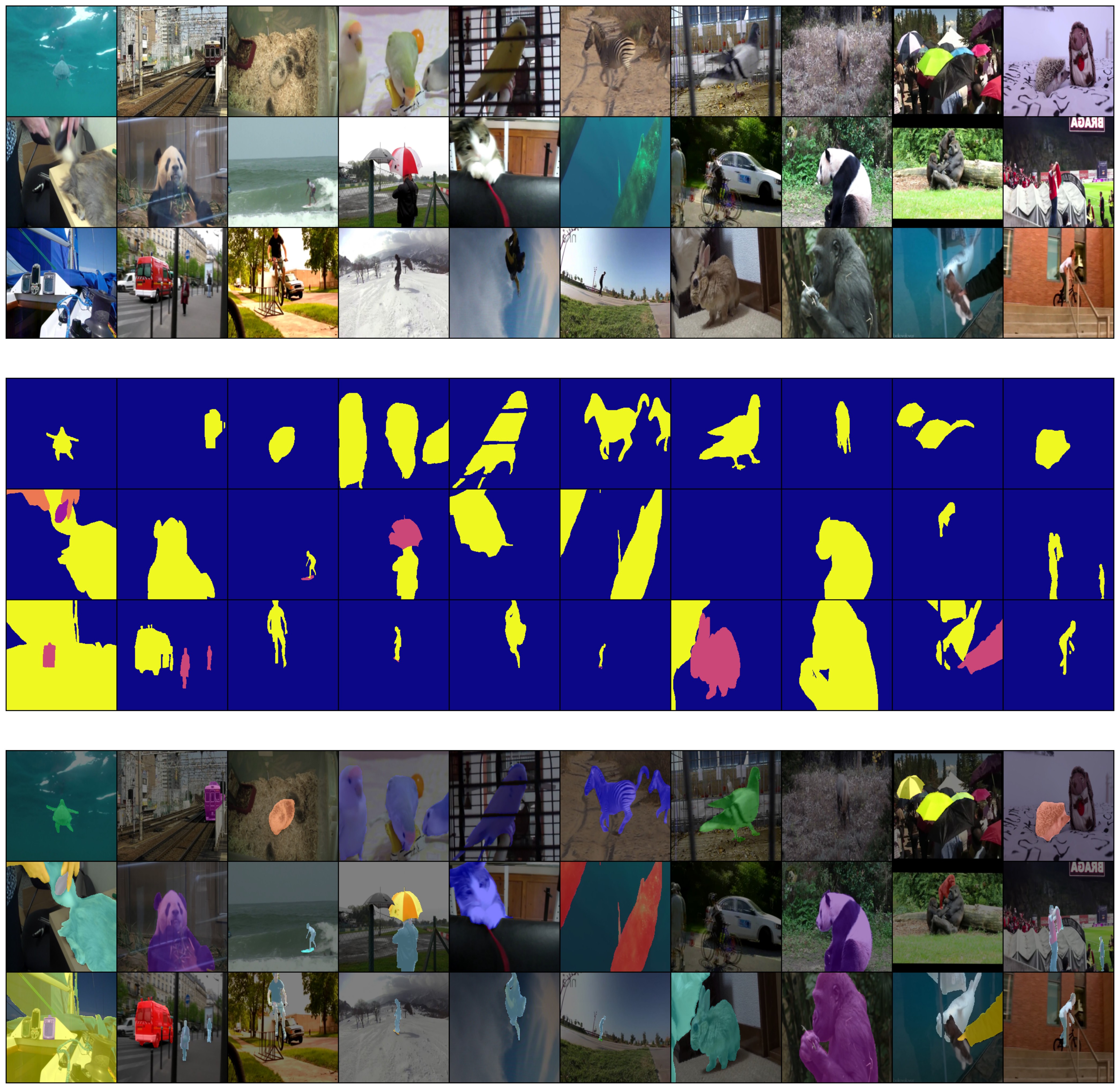}
  \caption{\textbf{YouTube VOS visualizations.} The first group of rows displays the images, the second shows the corresponding per-image masks, and the third overlays the masks on the images, aligned by their semantic IDs. These images and their ground truth segmentation maps are used for our tasks.}
  \label{fig:ytvos_mem}
\end{figure*}

\noindent \textbf{\textit{YTVOS}}~\cite{xu2018youtube} is another large-scale video object segmentation dataset, significantly larger than DAVIS17, comprising 4,453 videos annotated with 65 object categories. Similar to DAVIS17, ground-truth masks are provided only for the first frames of the test and validation sets. Therefore, a fixed 20\% subset of the training set is randomly sampled for evaluation; further details are provided in the supplementary material. Additionally, meta information is used to ensure that objects belonging to the same category retain consistent class IDs throughout the dataset, enabling semantic, category-level assessments. Figure~\ref{fig:appendix_class_distribution} illustrates the object distribution in YTVOS while \autoref{fig:ytvos_mem} showcases example frames and annotations.

\end{document}